\documentclass{article}
\usepackage{arxiv}
\usepackage[utf8]{inputenc}
\usepackage[pdftex]{graphicx}
\usepackage{mathtools}
\usepackage{amsmath}
\usepackage{array}
\usepackage{mathrsfs}
\usepackage{amssymb}
\usepackage{booktabs}
\usepackage{multirow}
\usepackage{tabularx}
\usepackage{tabulary}
\usepackage{dcolumn}
\usepackage{threeparttable}
\usepackage{bbm}
\usepackage{color, soul}
\usepackage{blindtext}
\usepackage{setspace}
\usepackage{geometry}
\usepackage{wrapfig}
\usepackage{lscape}
\usepackage{rotating}
\usepackage{epstopdf}
\usepackage[section]{placeins}
\usepackage{wasysym}
\usepackage[]{algorithm}
\usepackage{algpseudocode}
\usepackage{amsfonts}
\usepackage[toc,page]{appendix}
\usepackage{siunitx}
\usepackage{gensymb}
\usepackage{changes}
\usepackage{etoolbox}
\usepackage{multirow}
\usepackage{afterpage}
\usepackage{comment}
\usepackage{pgf}
\usepackage{siunitx}
\usepackage{standalone}
\usepackage{hyperref}
\usepackage[english]{babel}
\usepackage{fancyvrb}
\usepackage{esvect}

\hypersetup{
	colorlinks,
	linkcolor={blue!75!black},
	citecolor={blue!75!black},
	urlcolor={blue!75!black}
}
\providecommand{\keywords}[1]
{
	\small	
	\textbf{\textit{Keywords:}} #1
}

\allowdisplaybreaks
\usepackage{tikz}
\usetikzlibrary{shapes,arrows,fit,positioning,calc}
\usepackage[numbers,sort&compress]{natbib}
\tikzset{main node/.style={circle,draw=black,font=\sffamily\bfseries},edge_style/.style={draw=black,densely dashed},
}
\tikzset{
	>=stealth',
	punkt/.style={
		rectangle,
		rounded corners,
		draw=black, very thick,
		text width=8.5em,
		minimum height=2em,
		text centered},
	pil/.style={
		->,
		thick,
		shorten <=2pt,
		shorten >=2pt,}
}

\usepackage{metalogo}
\usepackage{dtklogos}
\allowdisplaybreaks 

\title{Simultaneous Decision Making for Stochastic Multi-echelon Inventory Optimization with Deep Neural Networks as Decision Makers}

\author{
	Mohammad Pirhooshyaran \\
	Industrial and Systems Engineering\\
	Lehigh University\\
	\texttt{mop216@lehigh.edu} \\
	\And 
	Lawrence V.\ Snyder \\
	Industrial and Systems Engineering\\
	Lehigh University\\
	\texttt{lvs2@lehigh.edu}	
}

\begin{document}
	
	\maketitle

	\begin{abstract}
		
		We propose a framework that uses deep neural networks (DNN) to optimize inventory decisions in complex multi-echelon supply chains. We first introduce pairwise modeling of general stochastic multi-echelon inventory optimization (SMEIO). Then, we present a framework which uses DNN agents to directly determine order-up-to levels between any adjacent pair of nodes in the supply chain. Our model considers a finite horizon and accounts for the initial inventory conditions. Our method is suitable for a wide variety of supply chain networks, including general topologies that may contain both assembly and distribution nodes, and systems with nonlinear cost structures. We first numerically demonstrate the effectiveness of the method by showing that its solutions are close to the optimal solutions for single-node and serial supply chain networks, for which exact methods are available. Then, we investigate more general supply chain networks and find that the proposed method performs better in terms of both objective function values and the number of interactions with the environment compared to alternate methods.
	\end{abstract}


	\section{Introduction}
\label{Intro}
The central goal in stochastic multi-echelon inventory optimization (SMEIO) is maintaining inventory levels by regulating the order quantities to optimize a cost function. The cost function usually consists of a shortage (penalty) cost plus holding costs, so that there is a tradeoff between ordering too much vs.~too little compared to the stochastic demand, each of which incurs its own cost.  In this study, we focus on identifying order-up-to levels (OULs) for multiple pairs of nodes in a complex supply chain network (SCN) under a finite decision horizon. Customer demands are stochastic with a known distribution; shipment lead-times are known constants; and decision makers have global information about the system states. 

To this end, we use the deep neural network (DNN) paradigm \cite{lecun2015deep} as a means for decision-making. We model general SCNs (we use the terms “mixed” and “general” SCN interchangeably) in which several DNNs can regulate separate inventory levels for different parts of the network jointly, and by interacting with other parts of the SCN environment, yet our model only suggests a order-up-to level for each decision.

Early approaches using machine learning (ML) to optimize inventory in SCNs primarily used finite Markov decision processes (MDP), dating to about two decades ago. \citet{giannoccaro2002inventory} introduce a simulation-based RL to optimize inventory decisions in a three-node serial SCN. Several modified Q-learning \cite{watkins1992q} algorithms have been presented, with Q-tables containing possible joint state--action spaces for SMEIO \cite{chaharsooghi2008reinforcement}.

Recently, \citet{oroojlooyjadid2017deep} explore the well-known beer game inventory problem (a four-node serial SCN) via the deep Q-learning framework \cite{mnih2013playing}, which integrates the DNN concept into a Q-learning algorithm. Previous ML studies are mainly focused on cost comparisons between their proposed approach and some known optimal or heuristic policies, and no further inventory policy behaviors have been reported \cite{chaharsooghi2008reinforcement,oroojlooyjadid2017deep}. Moreover, previous studies restrict the SMEIO settings to fit their assumptions. For instance, the action spaces (decisions made by ML approaches) are discretized in advance to make the problem tractable.

In this study, not only do we consider general complex SCNs with multiple decisions to be made at the same time by DNN agents at several echelons of the SCN, but we directly suggest OULs as the DNN's output. In other words, the proposed framework is capable of suggesting interpretable inventory actions. Moreover, there is no restriction on the order quantities (action space) and/or inventory levels (state space) of the proposed method. (For example, they are not discretized.)

As our first contribution, we model the SCNs considering pair-wise (edge) decision makers. Then, we propose a method that finds OULs for complex MEIO systems under a finite horizon. In order to demonstrate that our method is effective, we compare it against classical inventory optimization (IO) models and algorithms. However, these algorithms assume an infinite horizon. Therefore, to make the best comparison possible, we train our model in a finite-horizon setting that approaches steady-state sufficiently well that it is a reasonable approximation for the infinite-horizon setting. We do this by (a) setting the initial conditions carefully, (b) setting the decision horizon, and possibly the warm-up interval, carefully. For instance, we initialize the inventory levels to be lead-time demands. If our method is reasonably close to optimal for these classical IO settings, then we have confidence that it will also be close to optimal for settings that cannot currently be solved by classical IO models. The true value of our method is its ability to solve complex MEIO systems settings.

The rest of the paper is as follows: We briefly review the literature in Section \ref{Literature}, divided into two separate parts, one on SMEIO background and one on ML in the SMEIO framework. Then we explain our model in Section \ref{Model}. We describe our numerical experiments in Section \ref{Experiments} and we conclude the paper in Section \ref{Conclusion}. A python package containing an implementation of the framework is available at the paper's \href{https://github.com/mamadpierre/DNN-SMEIO}{repository}.  


\section{ML Background in SMEIO}
\label{Literature}
ML approaches for stochastic inventory optimization have been studied for a long time. Global supply chain management has been studied in \cite{pontrandolfo2002global} via the semi-Markov average reward technique (SMART). \citet{stockheim2003reinforcement} train RL agents to explore an optimal job acceptance strategy in a decentralized SCN. Many early works modify different versions of the Q-learning algorithm. Competitive supply chains are studied by \cite{van2007q}, where SCN nodes make their decisions separately and independently in an interconnected system.

\citet{oroojlooyjadid2017deep} study a four-node serial system via deep Q-learning \cite{mnih2013playing}. They  study cases in which (non-ML) nodes follow a base-stock policy, as well as cases in which those nodes display irrational behavior and diverge from their expected inventory policy. \citet{zhao2010application} investigate a multi-agent RL framework model to solve SMEIO considering multiple echelons and multiple commodities. \citet{chaharsooghi2008reinforcement} analyze supply chain ordering management with a focus on the beer game and suggest a reinforcement learning ordering mechanism. Recently, \citet{gijsbrechts2019can} extensively investigate the use of deep RL for three otherwise intractable inventory problems---dual sourcing, lost sales, and multi-echelon optimization problems.

Most previous studies consider discrete state and/or action spaces. A countable action set is a necessity for Q-learning convergence \cite{watkins1992q}; therefore, papers using Q-learning discretize the agent's possible action values. For instance, \citet{giannoccaro2002inventory} explain that in SMEIO, inventory position (IP) has no bound, which implies an infinite-size MDP. Then, they discretize the IP values and associate an integer number to an actual IP interval. New advances that integrate DNN into the RL framework open further opportunities to explore complex SCNs. There exist very few studies in which the policies are approximated by deep neural networks \cite{oroojlooyjadid2017deep}, but even these studies do not report the optimal base-stock levels or other inventory policy parameters. In other words, the ML framework aims to minimize the SCN cost function, but the appropriate interpretation of the solution into a near-optimal policy remains unexplored. We aim to provide a framework to present clear base-stock levels for general SCNs. An overview of SMEIO approaches is provided in Section \textcolor{blue}{1} of the supplementary.

\section {SMEIO Model and Environment}
\label{Model}

\subsection{SMEIO Model}

\begin{figure}
	\centering
\begin{tikzpicture}[shorten >=1pt, auto, node distance=3cm, thick,
   node_style/.style={circle,draw=black,font=\sffamily\Large\bfseries},
   node_style2/.style={circle,dashdotted,draw=black,font=\sffamily\Large\bfseries},
   edge_style/.style={draw=black,  thick},edge_style2/.style={draw=black,dashdotted,  thick},edge_style3/.style={draw=black,densely dashed,  thick}
   ]

    \node[node_style] (v1) at (-2,2) {2};
    \node[node_style] (v2) at (2,2) {4};
    \node[node_style2] (v3) at (4,2) {C};
    \node[node_style2] (v8) at (4,-2) {C};    
    \node[node_style] (v4) at (2,-2) {5};
    \node[node_style] (v5) at (-2,-2) {3};
    \node[node_style] (v6) at (-4,0) {1};
    \node[node_style2] (v7) at (-6.5,0) {$\infty$};    
    \draw[edge_style]  (v1) edge node{$LS_{24}=1$} (v2);
    \draw[edge_style2]  (v2) edge node{} (v3);
    \draw[edge_style]  (v4) edge node{$LS_{35}=1$} (v5);
    \draw[edge_style]  (v5) edge node{$LS_{13}=1$} (v6);
    \draw[edge_style]  (v4) edge node{$LS_{34}=1 \ \ \ $} (v1);    
    \draw[edge_style]  (v6) edge node{$LS_{12}=1$} (v1);
    \draw[edge_style2]  (v7) edge node{$LS_{\infty1}=2$} (v6);
    \draw[edge_style2]  (v4) edge node{} (v8);
    \draw[edge_style]  (v5) edge node{$LS_{25}=1 \ \ \ $} (v2);
    \end{tikzpicture}
	\caption{Mixed SCN.}
	\label{mixedSCN}
\end{figure}
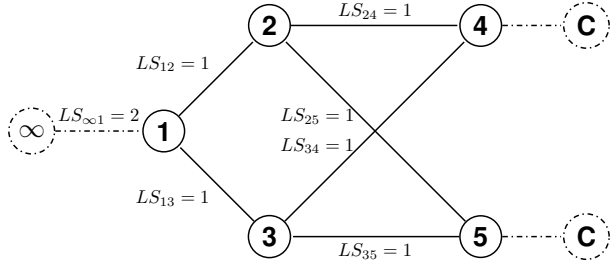

We consider a multi-echelon supply chain network with a general topology. The network must be connected and may not contain directed cycles; otherwise, any topology is allowed, including assembly nodes (nodes with more than one predecessor) and/or distribution nodes (nodes with more than one successor). We use $\mathcal{G}=(\mathcal{N}, \mathcal{E})$ to denote the SCN graph, in which $\mathcal{N}$ is the set of all nodes, and $\mathcal{E} \subseteq \{ (i,j): i,j \in \mathcal{N} \}$ is the set of all edges. We consider periodic review, with a finite horizon consisting of $T$ periods.
Demand at each customer-facing (leaf) node is stochastic and may have any probability distribution, discrete or continuous, so long as the demand is drawn iid from that distribution. Different leaf nodes may have different demand distributions. (Our example networks use normally distributed demands, except when stated otherwise.)  Figure \ref{mixedSCN} illustrates an example of an SCN, which we will return to later in the paper.

Each node in the network accepts raw materials from its supply node(s), processes them into finished goods, and ships the finished goods to its customer node(s). A given node's supply and customer nodes may be other nodes in the network (these are called {\em internal} supplier or customer nodes) or {\em external} suppliers or customers. We assume that a node cannot have {\em both} an internal {\em and} an external supplier, nor can it have both an internal and an external customer, but both of these assumptions can be relaxed through suitable use of dummy nodes. Let $\mathcal{U}_j$ be the set of immediate upstream nodes (i.e., predecessor nodes) from node $j$, and $\mathcal{D}_j$ be the set of immediate downstream nodes (successor nodes) from node $j$, for $j\in \mathcal{N}$. For modeling convenience, if a node has an external supplier or customer node, these are treated as dummy nodes in the network and are assumed to be contained in $\mathcal{U}_j$ and $\mathcal{D}_j$, respectively. Dummy supplier nodes are assumed to have infinite capacity. (The dummy supplier node for node 1 and the dummy customer nodes  for nodes 4 and 5 are indicated in Figure~\ref{mixedSCN} with dashed lines.)

Our framework allows for two different kinds of assembly nodes. An ``assembly-{\em and}'' node requires one unit of raw material from {\em each} of its predecessors---it is an ``and'' relationship. (It is straightforward to modify our approach to handle cases in which different numbers of units are required from different predecessors.) An ``assembly-{\em or}'' node requires one unit of raw material from {\em any} of its predecessors---it is an ``or'' relationship.

Each node has a finished-goods inventory that contains items that have been processed but not yet shipped to a customer. It also has one or more raw-material inventories, one for each supplier (including the external supplier, if any). When a node receives raw material items from its supplier(s), those items are placed into the raw material inventories. The node immediately processes as much raw material inventory as possible to produce finished goods. If node $j$ is an assembly-{\em and} node, then it processes
\begin{equation} R_{jt} = \min_{i\in \mathcal{U}_j} \{IL^r_{jit}\} \label{eq:Rjt_and} \end{equation}
items into finished goods in period $t$, and if it is an assembly-{\em or} node, then it processes
\begin{equation} R_{jt} = \sum_{i\in \mathcal{U}_j} \{IL^r_{jit}\} \label{eq:Rjt_or} \end{equation}
items, where $IL^r_{jit}$ is the number of units of the raw material from node $i$ that are in inventory at node $j$ in period $t$.  $R_{jt}$ units are immediately added to node $j$'s finished goods inventory. For an assembly-{\em and} node, $R_{jt}$ units are subtracted from each of node $j$'s raw material inventories. For an assembly-{\em or} node or a non-assembly node (i.e., a node with at most one predecessor), all units are removed from the raw material inventory.

The shipment lead time for orders placed by node $j$ from its predecessor $i\in \mathcal{U}_j$ is denoted $L_{ij}$. 
Shipment lead-times are deterministic, but due to possible upstream shortages, downstream nodes might experience stochastic lead-time. The processing time to convert raw materials to finished goods at a given node is assumed to be zero, though this assumption can be relaxed by adding dummy nodes whose shipment lead times equal the processing times.

If a node has insufficient inventory to meet its demands in a given period, the available inventory is allocated to customer nodes in proportion to the size of their orders in that period, and the remaining demands are backordered. Backorders are modeled as negative finished-goods inventory, as is common in the literature. If the inventory level is $IL$, then the number of items on hand is $IL^+$ and the number of backorders is $IL^-$, where $x^+ = \max\{x,0\}$; and $x^- = \max\{-x,0\}$.  Backorders may occur at any node, whether it has internal or external customers. However, since a node never processes more items than it has raw material inventory for, raw material inventories are always non-negative.

Holding and shortage costs may be arbitrary linear or nonlinear functions of the on-hand inventory (including in-transit inventory) and backorders, respectively, at the end of a period. 
In particular, the {\em state variables} at a given node $j\in \mathcal{N}$, evaluated at the end of period $t$, are as follows:
\begin{itemize}
	\item $IL_{jt}$ = the inventory level of finished goods 
	\item $BO_{jkt}$ = the backorders at node $j$ that are owed to customer node $k$ ($k\in \mathcal{D}_j$); note that $\sum_{k\in \mathcal{D}_k} BO_{jkt} = IL^-_{jt}$
	\item $IL^r_{jit}$ = the inventory level of raw material $i$ at node $j$ ($i\in \mathcal{U}_j$)
	\item $IT_{jkt}$ = the inventory in transit (being shipped) from node $j$ to node $k\in \mathcal{D}_j$
\end{itemize}
The {\em cost functions} at node $j$ are as follows: 
\begin{itemize}
	\item $h_{ij}(\cdot)$ = the holding cost function for items from node $i\in \mathcal{U}_j$ that are held in raw material inventory at $j$ or as a component of node $j$'s finished goods inventory or of in-transit inventory from node $j$ to node $k\in\mathcal{D}_j$
	\item $p_{jk}(\cdot)$ = the stockout cost function for backorders at node $j$ that are owed to node $k\in \mathcal{D}_j$
\end{itemize}
Then the total cost incurred in period $t$ is given by
\begin{equation} \label{eq:ct}
		c_t = \sum_{j\in \mathcal{N}}  \sum_{i\in \mathcal{U}_j} h_{ij}\left(IL^r_{jit} + IL^+_{jt} + \sum_{k\in \mathcal{D}_j} IT_{jkt}\right) 
		+ \sum_{j\in \mathcal{N}} \sum_{k\in \mathcal{D}_j} p_{jk}\left(BO_{jkt}\right) 
\end{equation}

where $h_{ij}(\cdot)$ and $p_{jk}(\cdot)$ are general (possibly non-linear or non-convex) functions. In this calculation, an item from node $i$ that is shipped to $j$ is counted in the holding cost function $h_{ij}(\cdot)$ when it is in raw-material inventory at node $j$, in finished-product inventory at node $j$, and in transit from node $j$ to its customer(s). This approach for calculating holding costs is somewhat non-standard, but is meant to provide more flexibility, and many common settings are special cases. For example, in a distribution system (each node has at most one predecessor) in which node $j$ has a holding cost charged on finished-goods inventory and inventory in transit to its successors, we can simply set $h_{ij}(x) = h_jx$ for all $(i,j)\in \mathcal{E}$. Moreover, we note that it is straightforward to modify our approach for cost functions with other functional forms. It is also worth mentioning that if one wants to consider an added holding cost value for finished items, one could model this by adding a dummy raw material.

At the end of the planning horizon (after period $T$ ends), any remaining inventory or backorders at node $j$ are reimbursed or charged according to a {\em salvage function} $v_j(x)$. That is, at the end of the horizon, the system incurs a cost of
\begin{equation} \sum_{j\in \mathcal{N}} v_j(IL_{jT}), \label{eq:salvage}  \end{equation}
where $v_j(x)$ may be positive (indicating a cost), negative (indicating a revenue), or zero, for either positive or negative values of $x$. Salvage values and costs are a common mechanism in inventory models to avoid end-of-horizon effects such as excess inventory buildup or selloffs near the end of the horizon. 

To facilitate traversing through the SCN graph, we number the nodes with integer values. We assume that the nodes are numbered $1,\ldots,N$ in ascending order by their total shipment lead-times from the infinite source. That is, if node $i$ has a longer total lead-time than $j$ as calculated from the infinite source, then $i > j$. Ties are broken arbitrarily, and duplicate node labels are not allowed.

\subsection{State Variables and Sequence of Events}
\label{sec:soe}

Each node $j$ follows a base-stock policy to place orders from its predecessors, and it may use a different order-up-to level (or base-stock level) for each  predecessor. In particular, we use $OUL_{ji}$ to represent the order-up-to level used by node $j\in \mathcal{N}$ when it places orders from predecessor $i\in\mathcal{U}_j$. The $OUL_{ji}$ values may be chosen by our DNN agent or by some other mechanism.

The sequence of events at each node $j$ in each time period $t$ is as follows:
\begin{enumerate}
	\item The demand $D_{jkt}$ is observed from each $k\in \mathcal{D}_j$. If $k$ is an internal customer, then $D_{jkt}$ is the order quantity placed by node $k$, and if $k$ is an external customer, then $D_{jkt}$ is an exogenous random variable.
	\item For each predecessor $i\in \mathcal{U}_j$, node $j$ orders $OUL_{ji} - IP_{jit}$ units from predecessor node $i$, where
	\begin{equation} IP_{jit} = IL_{j,i,t-1} - \sum_{k\in \mathcal{D}_j} D_{jkt} + IT_{i,j,t-1} + BO_{i,j,t-1} \label{eq:IP_def} \end{equation}
	is the inventory position of item-$i$ materials at node $j$ immediately before the order is placed. Note that it includes only the raw-material inventory of item $i$ at node $j$, and not the finished goods inventory at node $j$.
	\item For each predecessor $i\in \mathcal{U}_j$, node $j$ receives all items that were shipped from node $i$ $L_{ij}$ time periods ago. (There are $S_{i,j,t-L_{ij}}$ such units.) Raw material and in-transit inventories are updated as
	\begin{align*}
		IL^r_{jit} & = IL^r_{j,i,t-1} + S_{i,j,t-L_{ij}} \\
		IT_{ijt} & = IT_{i,j,t-1} - S_{i,j,t-L_{ij}}.
	\end{align*}
	\item Node $j$ processes $R_{jt}$ units, where $R_{jt}$ is given by \eqref{eq:Rjt_and} or \eqref{eq:Rjt_or} depending on whether node $j$ is an assembly-{\em and} node or an assembly-{\em or} node. (If $j$ is not an assembly node, then the two equations are equivalent.) The raw-material inventory levels are further updated as
	\begin{equation} IL^r_{jit} = IL^r_{jit} - R_{jt} \label{eq:ending_ILr_and} \end{equation} 
	if node $j$ is an assembly-{\em and} node, and as
	\begin{equation} IL^r_{jit} = 0 \label{eq:ending_ILr_or} \end{equation}
	otherwise. The finished goods inventory is updated as:
	\[ IL_{jt} = IL_{j,t-1} + R_{jt}. \]
	\item For each successor $k\in \mathcal{D}_j$, node $j$ ships $S_{jkt}$ units to node $k$. If $IL_{jt} \ge \sum_{k\in \mathcal{D}_j} (D_{jkt} + BO_{j,k,t-1})$, then node $j$ has sufficient inventory to meet all of its backorders and new demands; the shipment quantity and new backorder level are given by
	\begin{align}
		S_{jkt} & = D_{jkt} + BO_{j,k,t-1} \label{eq:S_enough} \\
		BO_{jkt} & = 0 \label{eq:ending_BO_enough}
	\end{align}
	If, instead, $IL_{jt} < \sum_{k\in \mathcal{D}_j} (D_{jkt} + BO_{j,k,t-1})$, then available inventory is allocated proportionally according to the current demands:
	\begin{align}
		S_{jkt} & = (D_{jkt} + BO_{j,k,t-1}) \frac{D_{jkt}}{\sum_{l\in \mathcal{D}_j} D_{jlt}} \label{eq:S_not_enough} \\
		BO_{jkt} & = BO_{j,k,t-1} - S_{jkt}. \label{eq:ending_BO_not_enough}
	\end{align}
	In either case, the finished-goods and in-transit inventory levels are updated as
	\begin{align}
		IL_{jt} & = IL_{jt} - \sum_{k\in \mathcal{D}_j} D_{jkt} \label{eq:ending_IL} \\
		IT_{jkt} & = IT_{jkt} + S_{jkt}. \label{eq:ending_IT}
	\end{align}
	(Note that only new demands, not old backorders, are subtracted from $IL_{jt}$ since old backorders are already counted as negative inventory in $IL_{jt}$. Note also that demands are subtracted whether or not they are actually shipped out, since the inventory level decreases in either case, either by a reduction in on-hand inventory or an increase in backorders.)
	\item Holding and stockout costs are assessed according to \eqref{eq:ct}.
	The ending raw-material inventory levels are given by \eqref{eq:ending_ILr_and} or \eqref{eq:ending_ILr_or}; the ending finished-goods inventory level is given by \eqref{eq:ending_IL}; the ending backorders are given by \eqref{eq:ending_BO_enough} or \eqref{eq:ending_BO_not_enough}; and the ending in-transit inventories are given by \eqref{eq:ending_IT}.
\end{enumerate}
The sequence described above is the sequence of events for each node. However, the events are split into two phases: In the first phase, the nodes follow events 1--3 in order from downstream to upstream, and in the second phase, the nodes follow events 4--6 in order from upstream to downstream. That is, the downstream-most nodes place their orders to their predecessors, their predecessors place their orders, etc.; then the upstream-most nodes ship units to their successors, who ship units to their successors, etc.

Note also that the multi-period newsvendor problem (in which there is a single node, which can hold inventory and backorders from one period to the next) can be modeled using the framework above by setting the lead time to 1. (The newsvendor problem is often described as having zero lead time, but it also uses a different sequence of events, in which we observe the demand after we place the order. Setting the lead time to 1 converts our sequence of events to this one.)

The notation is summarized in Section \textcolor{blue}{2} of supplementary material.
\subsection{Interaction between Agents and Environment}
\label{sec:interaction}

\begin{figure}
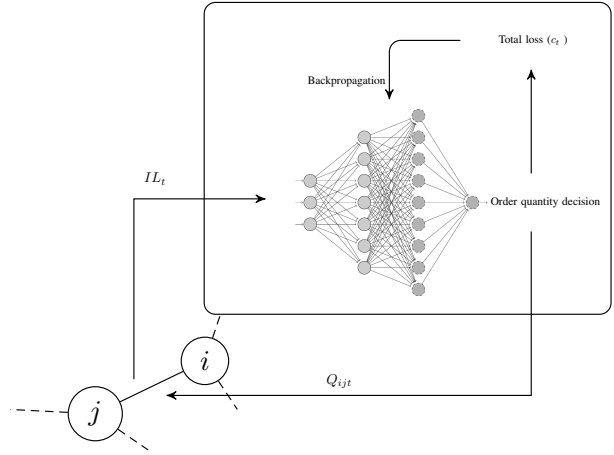

	\hspace*{-1cm}
	\centering
	\includestandalone[width = 8cm]{frameworkDNN}
	\caption{DNN-SMEIO framework.}
	\label{framework}
\end{figure}
Each node $j$ must choose the order-up-to level (OUL) it uses when placing order from each of its predecessor nodes $i\in \mathcal{U}_j$; we refer to the OULs as being chosen ``for the edge $(i,j)$.'' Decisions for each edge $(i,j)$ may be made by a separate DNN agent, or using some exogenous mechanism. For ease of exposition, we will assume that {\em all} edges are governed by a DNN agent, but it is straightforward to adapt our method if some edges have non-DNN decision makers. This structure implies that there can be up to $|\mathcal{E}|$ independent DNN decision makers.

The agents are trained by interacting with an {\em environment} consisting of a simulation of the SCN described above. In particular, the agent makes decisions for multiple {\em episodes}, each of which consists of $T$ time periods. At the beginning of each episode $k$, the DNN chooses order-up-to levels $OUL_{ji}$ for all $(i,j)\in \mathcal{E}$ and sends these levels to the environment. The environment simulates the SCN to calculate the cost of the current OULs, given by 
\begin{equation}\label{eq:ck}
	\mathcal{C}_k = \sum_{t=1}^{T} c_t, 
\end{equation}
where $c_t$ is as given by equation \eqref{eq:ct}. The DNN weights are then updated, new OULs are chosen, and a new episode begins. When the DNNs are trained, the output $OUL_{ji}$ converges to a single OUL. In practice, the weight-update procedure happens considering mini-batches of episodes together for computational purposes. The inputs to the DNN are largely irrelevant and can be set in any number of ways. This is because the DNN is optimizing an objective function, rather than trying to determine labels for a given input. In our numerical experiments, we use the inventory positions as the input, but the DNN could just as easily be given a vector of 1s as its input.

Figure \ref{framework} shows  the proposed framework, in which a DNN is responsible for deciding the quantities of items ordered by node $j$ from node $i$. We use fully connected DNNs, with several possible hidden layers for each decision maker. 
Moreover, a batch normalization procedure is considered after every network layer, which considerably stabilizes the learning procedure.

We allow the agents to be optimized by individual DNNs or by a single DNN. That is, networks corresponding to different agents can be completely separate, or they can share their first few layers.

The loss function used to train the DNN(s) is given by the per-episode costs $\mathcal{C}_k$ given in \eqref{eq:ck}. (A similar approach was used by \cite{oroojlooyjadid2020applying}. This is in contrast to the more common approach in which the DNN loss function measures the distance between an estimated and actual value.) To optimize the weights of the DNN network(s), we use the adaptive moment estimation (Adam) optimizer \cite{kingma2014adam} with fixed learning rate. Each environment consists of many hyperparameters that might need to be tuned. Details on framework structure, optimization algorithm is provided in Section \textcolor{blue}{3} of supplementary material.

\section{Numerical Experiments}
\label{Experiments}
We report the results of our experiments on single-node, serial, assembly and mixed SCNs. At first, we investigate the effectiveness of the model against infinite-horizon classical IO models to check whether our proposed method can produce results that are close to those produced by established methods for classical IO problems. We utilize Spearmint Bayesian optimization, and in particular, the Gaussian process expected improvement (GPEI) method for hyperparameter tuning of our framework \cite{swersky2013multi} (Refer to Section \textcolor{blue}{3} of supplementary material).

\subsection{Single-Node SCN}

\begin{table}[]
	\caption{OUL and cost comparison for single-node inventory instances}
	\label{tab1}
	\setlength\tabcolsep{1pt}
	\centering	
	\begin{tabular}{ccccccccccccccccccc}
		&  &                &  & \multicolumn{7}{c}{$L=0$}                          &  & \multicolumn{7}{c}{$L=1$}                           \\ 
		& & demand                       &  & \multicolumn{3}{c}{analytical} &  & \multicolumn{3}{c}{DNN} &  & \multicolumn{3}{c}{analytical} &  & \multicolumn{3}{c}{DNN} \\ \cline{5-11} \cline{13-19}
		case \# & & distribution                       &  & OUL        &       & cost      &  & OUL      &    & cost    &  & OUL          &    & cost       &  & OUL       &  & cost     \\ 
		\hline
		1       &  & $\mathcal{N}(10,1)$   &  & $ 0 $      &       & $ 0 $     &  & $ 0 $    &    & $ 0 $   &  & $10.67  $    &    & $12.71$    &  & $10.68$   &  & $12.71$  \\
		2       &  & $\mathcal{N}(10,2)$   &  & $ 0 $      &       & $ 0 $     &  & $ 0 $    &    & $ 0 $   &  & $11.35$      &    & $25.42$    &  & $11.50$   &  & $25.47$  \\
		3       &  & $\mathcal{N}(50,1)$   &  & $ 0 $      &       & $ 0 $     &  & $ 0 $    &    & $ 0 $   &  & $50.67$      &    & $12.71$    &  & $50.58$   &  & $12.75$  \\
		4       &  & $\mathcal{N}(50,5)$   &  & $ 0 $      &       & $ 0 $     &  & $ 0 $    &    & $ 0 $   &  & $ 53.37$     &    & $63.56$    &  & $ 53.30$  &  & $63.59$  \\
		5       &  & $\mathcal{N}(100,1)$  &  & $ 0 $      &       & $ 0 $     &  & $ 0 $    &    & $ 0 $   &  & $100.67$     &    & $12.71$    &  & $100.77$  &  & $12.75$  \\
		6       &  & $\mathcal{N}(100,5)$  &  & $ 0 $      &       & $ 0 $     &  & $ 0 $    &    & $ 0 $   &  & $103.37$     &    & $63.56$    &  & $103.28$  &  & $63.58$  \\
		7       &  & $\mathcal{N}(100,10)$ &  & $ 0 $      &       & $ 0 $     &  & $ 0 $    &    & $ 0 $   &  & $106.74$     &    & $127.11$   &  & $106.79$  &  & $127.12$ \\ \hline
	\end{tabular}
\end{table}

In this section, we consider a simple SCN consisting of a single node. In particular, the node has holding and stockout costs of $h=10$ and $p=30$, respectively. The demand per period is normally distributed; we consider various means and standard deviations. We set the holding cost $h=10$ and shortage cost $p=30$ for all the cases. This means critical ratio equals $0.75$. We consider a short time horizon of $T=2$ periods, with no salvage value ($v(x) = 0$ for all $x$). We consider two settings for the lead time, one in which $L=1$ and one in which $L=0$. In the $L=0$ case, the decision maker sees the demand and then places an order; there is no stochasticity. The optimal action is simply to order the realized demand value, and the optimal cost is 0. This is not a typical setting, but we examine it to evaluate the learning process of the DNN-SMEIO framework in a very simple case. 

The $L=1$ case is mathematically equivalent to the newsvendor problem. (Recall that in the classical newsvendor problem, the order is received before the demand is observed, whereas the opposite is true in our sequence of events. Therefore, our model is equivalent to the newsvendor model if we set $L=1$.)

Table \ref{tab1} reports the 7 instances we tested. The table shows the demand distribution, the optimal OUL and average cost per period, and the OUL and average cost given by the DNN, for both the $L=0$ and $L=1$ cases. The optimal OULs were found using an analytical approach. For the DNN solutions, we calculate the cost using the base-stock policy simulation.
One can see that the results are very close; the DNN finds near-optimal OULs for these instances. The largest relative error between DNN-SMEIO approach and the analytical solutions are 1.32\% and 0.31\% for the OUL and cost values, respectively.

\begin{figure*}
	\hspace*{-1.5cm}
	\scalebox{0.4}{\input{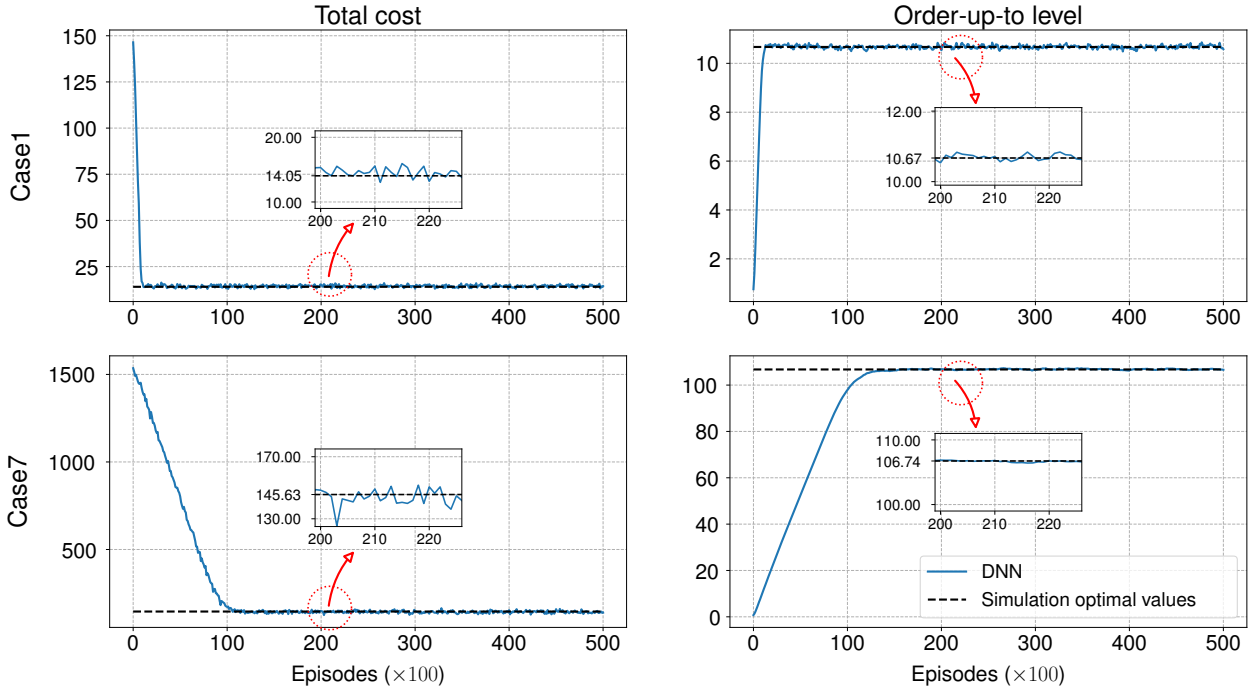}}
	\caption{Loss function and OUL convergence curves for single-node inventory instances case 1 and 7.}
	\label{fig:newsvendorConvergenceCurve}
\end{figure*}

Figure \ref{fig:newsvendorConvergenceCurve} illustrates the loss function and OUL convergence curves for instances 1 and 7 from Table \ref{tab1} and for both lead-time settings. We train the DNN-SMEIO for 50000 training episodes. The x-axes contain 500 points. After each 100 episodes of training, a new set of episodes is conducted for testing. The cost and OULs are calculated, which corresponds to a single point in the respective figures. One can stop the training much sooner than 50000 episodes. For instance, case 1 reaches its best result after only 1500 training episodes. We let the process continue only to demonstrate the method's stability. The black dotted lines for the OULs are the true optimal values. However, for the loss function, we simulate the process identically to the one that DNN-SMEIO is trained over (i.e., we consider the same initialization and episode horizon). 

\subsection{Serial SCN}
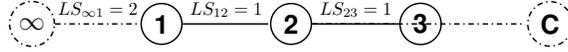
\begin{figure}
	\centering
\begin{tikzpicture}[shorten >=1pt, auto, node distance=3cm, thick,
   node_style/.style={circle,draw=black,font=\sffamily\Large\bfseries},
   node_style2/.style={circle,dashdotted,draw=black,font=\sffamily\Large\bfseries},
   edge_style/.style={draw=black,  thick},edge_style2/.style={draw=black,dashdotted,  thick},edge_style3/.style={draw=black,densely dashed,  thick}
   ]

    \node[node_style] (v1) at (-1.5,0) {2};
    \node[node_style2] (v3) at (3.5,0) {C};
    \node[node_style] (v5) at (1,0) {3};
    \node[node_style] (v6) at (-4,0) {1};
    \node[node_style2] (v7) at (-6.5,0) {$\infty$};  
    
    \draw[edge_style2]  (v7) edge node{$LS_{\infty1}=2$} (v6);
    \draw[edge_style]  (v6) edge node{$LS_{12}=1$} (v1);    
    \draw[edge_style]  (v1) edge node{$LS_{23}=1$} (v5);

    \draw[edge_style2]  (v1) edge node{} (v3);

    \end{tikzpicture}
	\caption{Three-echelon serial SCN.}
	\label{serialSCN}
\end{figure}

In this section, we consider instances of serial SCNs. First, we discuss the comparison structure through a three-echelon serial system, which is illustrated in Figure \ref{serialSCN} and used as an example by \citet{fosct2e}. The network has 3 nodes and therefore 3 separate decisions to be made. There is a single external customer with normally distributed demand $\mathcal{N}(5,1)$. The (local) holding costs increase as one moves downstream, and there is a shortage cost only at the furthest downstream node. \citet{fosct2e} report the optimal OULs for this system under an infinite-horizon, continuous-review environment, based on the Clark--Scarf recursive method  \cite{clark1960optimal,ChenZheng94}. However, as mentioned previously, our environment is a finite-horizon, periodic-review environment. Although the two environments are not strictly comparable, our time horizon (we use $T=10$) is long enough so that the system approximately reaches steady state and can therefore be compared to the infinite-horizon case.
Moreover, after the agents report their OULs, we use a simple base-stock policy simulation to find the cost. Nevertheless, because of the inherent differences between our assumptions (finite horizon, periodic review, cost evaluation by simulation) and those of the Clark-Scarf method (infinite horizon, continuous review, analytical cost evaluation), we also consider a third approach for optimizing the OULs: derivative-free optimization (DFO). We use the Trust Region DFO (DFO-TR) method \cite{bandeira2012computation}, which is a model-based DFO method and is well-known to use as few function evaluations as possible. DFO can be used to explore the best possible OULs, so we use them as an additional benchmark against which to compare the DNN-SMEIO method.

The methods see the optimal OULs neither at the initialization nor during the training process. The problem settings are reported in Table \ref{tab31}, and the analytical, DNN-SMEIO and DFO results are presented in Table \ref{tab32}. As one can see, the DNN results are quite close to the ones suggested by the analytical approach. For instance, the difference between the best cost for the case discussed above (case 3) achieved by the proposed framework ($47.90$) and the cost obtained by simulation using the exact OULs ($47.65$) is less than 1\%. DFO gives considerably worse results for cases 7 and 8 but performs as well as the other approaches for the rest of the cases.

\begin{figure*}
	\hspace*{-1.25cm}
	\scalebox{0.40}{\input{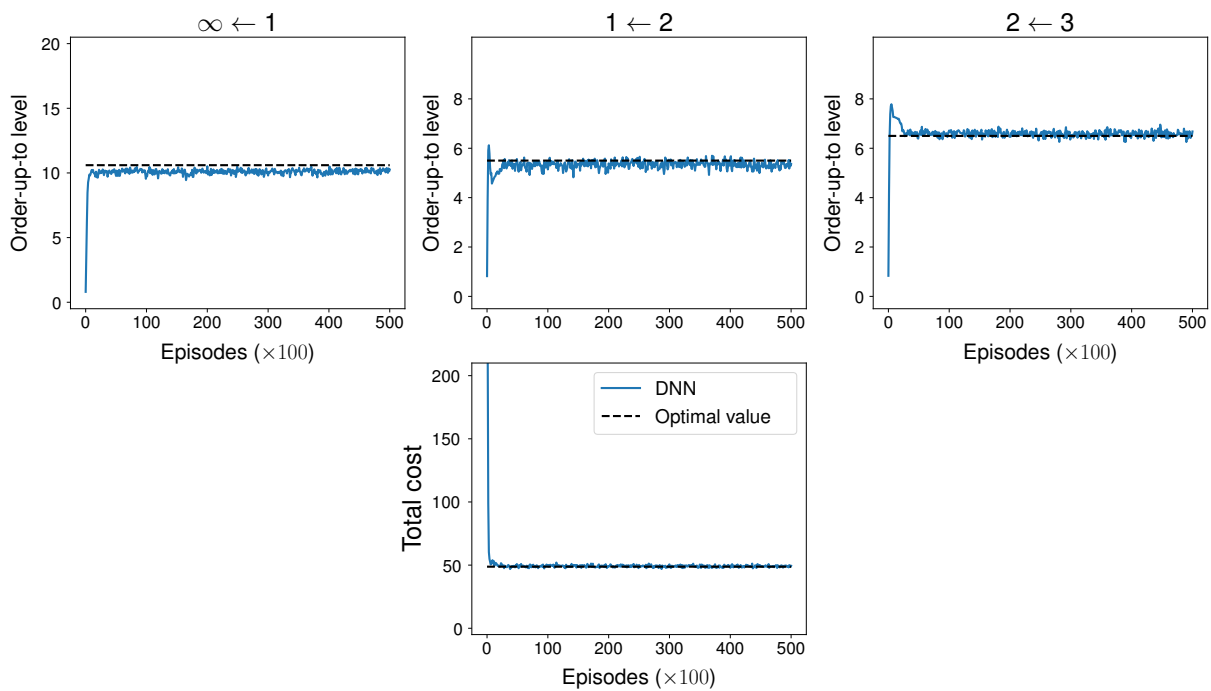}}
	\caption{Total cost and OUL decisions convergence curves for serial SCN case 3}
	\label{fig:THreeEchelonSMEIO}
\end{figure*}

The convergence curves for the loss function and the OULs of serial SCN case 3 are provided in Figure \ref{fig:THreeEchelonSMEIO}. Training and testing procedures are similar to those discussed in the previous section. One can see that the DNN-SMEIO method is stable to a great extent. 
\begin{table*}[]
	\caption{Serial SCNs settings}
	\label{tab31}
	\setlength\tabcolsep{2pt}
	\centering	
	\begin{tabular}{ccccccccccc}
		case &&	\# echelons &  & demands            &  & holding cost         &  & shortage cost            &  & lead-time       \\     
		&&	&  &                    &  & (per item)           &  & (per item)               &  &                          \\ 
		\hline
		1&&	2         &  & $\mathcal{N}(3,0.5)$ &  & $\left(5,8.2\right)$ &  & $\left(0,25.5\right)$ &  & $\left(1,1\right)$ \\
		
		2&&	2         &  & $\mathcal{N}(6,1.5)$ &  & $\left(1.9,4.1\right)$ &  & $\left(0,11.3\right)$ &  & $\left(2,1\right)$  \\	
		
		3&&	3         &  & $\mathcal{N}(5,1)$ &  & $\left(2,4,7\right)$ &  & $\left(0,0,37.12\right)$ &  & $\left(2,1,1\right)$ \\
		4&&	3         &  & $\mathcal{N}(50,3)$ &  & $\left(5,10,25\right)$ &  & $\left(0,0,50\right)$ &  & $\left(2,1,1\right)$  \\
		5&&	3         &  & $\mathcal{N}(100,5)$ &  & $\left(25,25,50\right)$ &  & $\left(0,0,100\right)$ &  & $\left(1,2,2\right)$ \\
		6&&	3         &  & $\mathcal{N}(100,10)$ &  & $\left(10,20,30\right)$ &  & $\left(0,0,100\right)$ &  & $\left(1,1,1\right)$  \\
		
		7&&	4         &  & $\mathcal{N}(3,0.4)$ &  & $\left(4,5.75,7.90,10.8\right)$ &  & $\left(0,0,0,35.5\right)$ &  & $\left(1,1,1,1\right)$  \\
		
		8&&	4         &  & $\mathcal{N}(5,1.2)$ &  & $\left(5,5,5,10\right)$ &  & $\left(0,0,0,30\right)$ &  & $\left(1,1,1,1\right)$  \\
		9&&	5         &  & $\mathcal{N}(80,4)$ &  & $\left(10,20,30,40,50\right)$ &  & $\left(0,0,0,0,200\right)$ &  & $\left(1,1,1,1,1\right)$  \\
		10&&	5         &  & $\mathcal{N}(25,2)$ &  & $\left(5,10,25,50,50\right)$ &  & $\left(0,0,0,0,150\right)$ &  & $\left(2,1,1,1,1\right)$  \\				
		\hline
	\end{tabular}
\end{table*}

\begin{table*}[]
	\caption{OUL and cost comparison for serial SCNs}
	\label{tab32}
	\setlength\tabcolsep{2pt}
	\hspace*{-1.5cm}
	\small
	\begin{tabular}{ccccccccccccc}
		&  & \multicolumn{3}{c}{Analytical}     &  & \multicolumn{3}{c}{DNN}     &&  \multicolumn{3}{c}{DFO}    \\  
		\cline{3-5} \cline{7-9} \cline{11-13} 
		case    &  & OULs                  &  & cost    &  & OULs                &  & cost  &  & OULs                &  & cost  \\ 
		\hline
		1 &  & $(2.91, 3.64)$ &  & $22.21$ &  & $(2.91, 3.72)$ &  & $22.34$ && $(1.22, 5.10)$ && $22.55$\\
		2 &  & $(12.58,7.60)$ &  & $23.07$ &  & $(12.58,7.65)$ &  & $23.17$ && $(12.05,  7.58) $&&  $23.20$\\	
		3 &  & $(10.69,5.53,6.49)  $ &  & $47.65$ &  & $(10.08,5.39,6.64)$ &  & $47.90$  && $(10.54,  5.35,  6.57)$ && $50.01$\\
		4 &  & $(101.45,51.40,52.7040)  $ &  & $879.88$ &  & $(99.29,51.03,52.71)$ &  & $885.63$ && $(97.02, 53.59, 53.02)$ && $885.49$ \\
		5 &  & $(71.026, 228.29,  207.04)  $ &  & $10568.23$ &  & $(87.71,204.63,208.90)$ &  & $10625.01$ && $(79.33, 211.20, 208.51)$ && $10695.88$ \\
		6 &  & $(99.53,102.58,114.05)  $ &  & $3630.14$ &  & $(93.36,103.42,114.26)$ &  & $3651.63$ && $(95.83, 100.87, 117.90)$&& $3638.18$\\
		7 &  & $(2.78,3.13,3.19,3.60)  $ &  & $63.39$ &  & $(2.78, 3.13, 3.19, 3.74)$ &  & $63.84$ && $(-12.03,  -8.86,  -3.93,   1.90)$ && $592.51$\\
		8 &  & $(-3.80,9.80,9.80,6.35)  $ &  & $101.48$ &  & $(1.48,6.12,7.00,6.46)$ &  & $104.04$ && $(-4.96, -3.69, -1.90,  0.02)$&& $674.62$ \\
		9 &  & $(80.15,80.15,81.17,81.68,86.99)  $ &  & $8559.85$ &  & $(76.83,78.02,79.60,81.62,87.40)$ &  & $8678.38$ && $(80.49, 77.62, 80.04, 77.45, 92.18)$ && $8585.50$\\
		10 &  & $(51.57,   26.30,   25.05,   20.25,   33.01)$ &  & $2500.79 $ &  & $(48.40,   25.65,   24.02,   22.90,   30.12)$ &  & $2581.41 $ && $(49.44, 25.51, 23.04, 23.82, 31.17)$ && $2527.1$ \\						
		\hline
	\end{tabular}
\end{table*}

\subsection{Assembly SCNs}
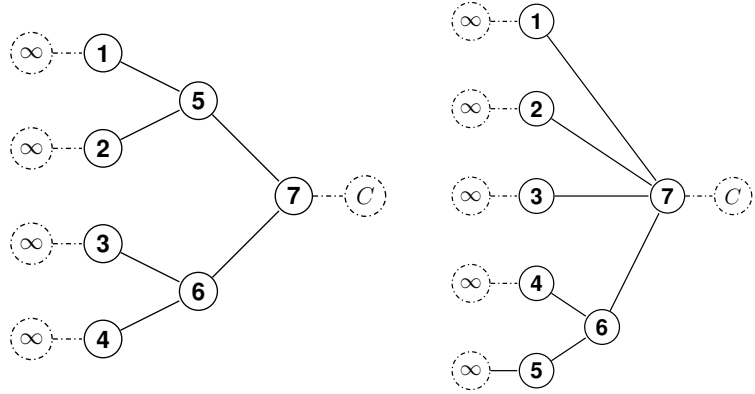
\begin{figure*}
	\centering
	\begin{minipage}{.35\textwidth}
			
	\begin{tikzpicture}[shorten >=1pt, auto, node distance=3cm, thick,
		node_style/.style={circle,draw=black,font=\sffamily\Large\bfseries},
		node_style2/.style={circle,dashdotted,draw=black,font=\sffamily\Large\bfseries},
		edge_style/.style={draw=black,  thick},edge_style2/.style={draw=black,dashdotted,  thick},edge_style3/.style={draw=black,densely dashed,  thick}
		]

		\node[node_style2] (v1) at (-6.5,3) {$\infty$};    
		\node[node_style2] (v2) at (-6.5,1) {$\infty$};    
		\node[node_style2] (v3) at (-6.5,-1) {$\infty$};    
		\node[node_style2] (v4) at (-6.5,-3) {$\infty$};    
		\node[node_style]  (v5) at (-5,3) {1};
		\node[node_style]  (v6) at (-5,1) {2};
		\node[node_style]  (v7) at (-5,-1) {3};
		\node[node_style]  (v8) at (-5,-3) {4};    
		\node[node_style]  (v9) at (-3,2) {5};
		\node[node_style]  (v10) at (-3,-2) {6};
		\node[node_style]  (v11) at (-1,0) {7};
		\node[node_style2] (v12) at (0.5,0) {$C$};
		
		\draw[edge_style2]  (v1) edge node{} (v5);
		\draw[edge_style2]  (v2) edge node{} (v6);
		\draw[edge_style2]  (v3) edge node{} (v7);
		\draw[edge_style2]  (v4) edge node{} (v8);
		\draw[edge_style]   (v5) edge node{} (v9);    
		\draw[edge_style]   (v6) edge node{} (v9);    
		\draw[edge_style]   (v7) edge node{} (v10);
		\draw[edge_style]   (v8) edge node{} (v10);
		\draw[edge_style]   (v9) edge node{} (v11);
		\draw[edge_style]   (v10) edge node{} (v11);
		\draw[edge_style2]  (v11) edge node{} (v12);
	\end{tikzpicture}
	\end{minipage}
	\begin{minipage}{.15\textwidth}
			
	\begin{tikzpicture}[shorten >=1pt, auto, node distance=3cm, thick,
		node_style/.style={circle,draw=black,font=\sffamily\Large\bfseries},
		node_style2/.style={circle,dashdotted,draw=black,font=\sffamily\Large\bfseries},
		edge_style/.style={draw=black,  thick},edge_style2/.style={draw=black,dashdotted,  thick},edge_style3/.style={draw=black,densely dashed,  thick}
		]

		\node[node_style2] (v1) at (-6.5,4) {$\infty$};    
		\node[node_style2] (v2) at (-6.5,2) {$\infty$};    
		\node[node_style2] (v3) at (-6.5,0) {$\infty$};    
		\node[node_style2] (v4) at (-6.5,-2) {$\infty$}; 
		\node[node_style2] (v5) at (-6.5,-4) {$\infty$}; 
		
		\node[node_style]  (v6) at (-5,4) {1};
		\node[node_style]  (v7) at (-5,2) {2};				   
		\node[node_style]  (v8) at (-5,0) {3};
		\node[node_style]  (v9) at (-5,-2) {4};		
		\node[node_style]  (v10) at (-5,-4) {5};		

		\node[node_style]  (v11) at (-3.5,-3) {6};		
		\node[node_style]  (v12) at (-2,0) {7};						
		
		\node[node_style2]  (v13) at (-0.5,0) {$C$};

		\draw[edge_style2]  (v1) edge node{} (v6);
		\draw[edge_style2]  (v2) edge node{} (v7);
		\draw[edge_style2]  (v3) edge node{} (v8);
		\draw[edge_style2]  (v4) edge node{} (v9);
		\draw[edge_style]   (v5) edge node{} (v10);    
		\draw[edge_style]   (v9) edge node{} (v11);    
		\draw[edge_style]   (v10) edge node{} (v11);
		\draw[edge_style]   (v6) edge node{} (v12);
		\draw[edge_style]   (v7) edge node{} (v12);
		\draw[edge_style]   (v8) edge node{} (v12);
		\draw[edge_style]  (v11) edge node{} (v12);
		\draw[edge_style2]  (v12) edge node{} (v13);

	\end{tikzpicture}
	\end{minipage}
	\caption{Assembly SCN structures. Left figure: assembly SCN (1) and the right figure assembly SCN (2).}
	\label{AssemblySCN}			
\end{figure*}

We consider two separate assembly structures, which are illustrated in Figure \ref{AssemblySCN}. Each structure contains three echelons. There are 10 and 11 OUL decisions to be made for assembly SCNs 1 and 2, respectively. For each of these structures, we consider 5 separate cases having different holding and shortage costs, lead-times and demand distributions. We use DFO, coordinate descent (CD), and enumeration as heuristic benchmarks to compare our method against. For CD and enumeration, each candidate solution is evaluated by simulating the system for 3 trials, each consisting of 200 periods; one solution is considered superior to another if the mean of the 3 total costs (one per trial) for that solution is smaller than that of the other. Lower and upper bounds for both CD and enumeration are set equal to $0.75D$ and $2D$, respectively, where $D$ is the mean lead-time demand observed by the node. For enumeration, the resulting range was discretized into 10 equal intervals. Moreover, we restricted the OULs to be equal at all nodes within a given echelon, which is optimal given the symmetries in the SCNs, and which therefore reduces the search space. Note that we did not make the same restriction for the DNN method, meaning that we are providing an advantage to the benchmark methods. The complete specifications of the parameters of the assembly system instances, as well as the OULs suggested by each method, are presented in Tables \textcolor{blue}{3} and \textcolor{blue}{4} of the supplement. 

For all solutions produced by all methods, we evaluated the cost by simulating the system for 10 trials, each consisting of 10,000 periods. The mean of the 10 total costs (one per trial) is reported in the  Table \ref{tabAssembly}. We observe that the costs of the solutions suggested by DNN are very close to those from CD and enumeration, even though the search space was restricted as described above for CD and enumeration but not for DNN. 

DFO is inferior compared to the other approaches. For four of the cases, the DFO OULs did not converge to reasonable values, resulting in a very large cost. There are at least three possible reasons for this. First, we may have exceeded the number of variables that can be properly handled by DFO. Second, DFO methods are generally local methods, and DFO may become trapped in local minimum. Third, the performance of DFO heavily depends on the starting point, and we may have started from OULs that are far from the optimal ones, although we tried to alleviate its effect by starting with lead-time demands as initial inventory levels.

\begin{table}[]
	\caption{Cost comparisons for assembly SCNs}
	\label{tabAssembly}
	\setlength\tabcolsep{1pt}
	\centering	
	\begin{tabular}{cccccc}
		& Case &DNN-SMIO  & CD & enumeration & DFO\\ 
		\hline
		assembly1 & 1 & 40.55 & 40.27  & 40.34    &233.45 \\\cline{1-1}
		&2 & 103.77  & 101.59            & 101.47   &482.63  \\
		&3 & 163.15  & 161.30            & 161.13   &441.43 \\
		&4 & 37.49  & 35.97            & 35.98     &139.77 \\
		&5 & 29.04  & 27.53            & 27.45      &36.03 \\ \hline
		average  &   & 74.80 & 73.33 & 73.27         &324.32  \\
		\hline
		&        & &                    &        &      \\
		assembly2 & 1 & 93.94 & 90.40    & 90.54    &116.38 \\\cline{1-1}
		&2 & 23.00    & 22.43            & 22.48      &25.75 \\
		&3& 86.61    & 82.67            & 82.32      &90.71 \\
		&4& 34.62 & 34.04            & 34.17      &42.35 \\
		&5& 30.98 & 28.19            & 27.96      &62.76 \\   \hline
		average & & 53.83 & 51.55 & 51.49 & 67.59
	\end{tabular}
\end{table}

\subsection{Mixed SCN}

\begin{figure*}
	\hspace*{-1.25cm}
	\scalebox{0.40}{\input{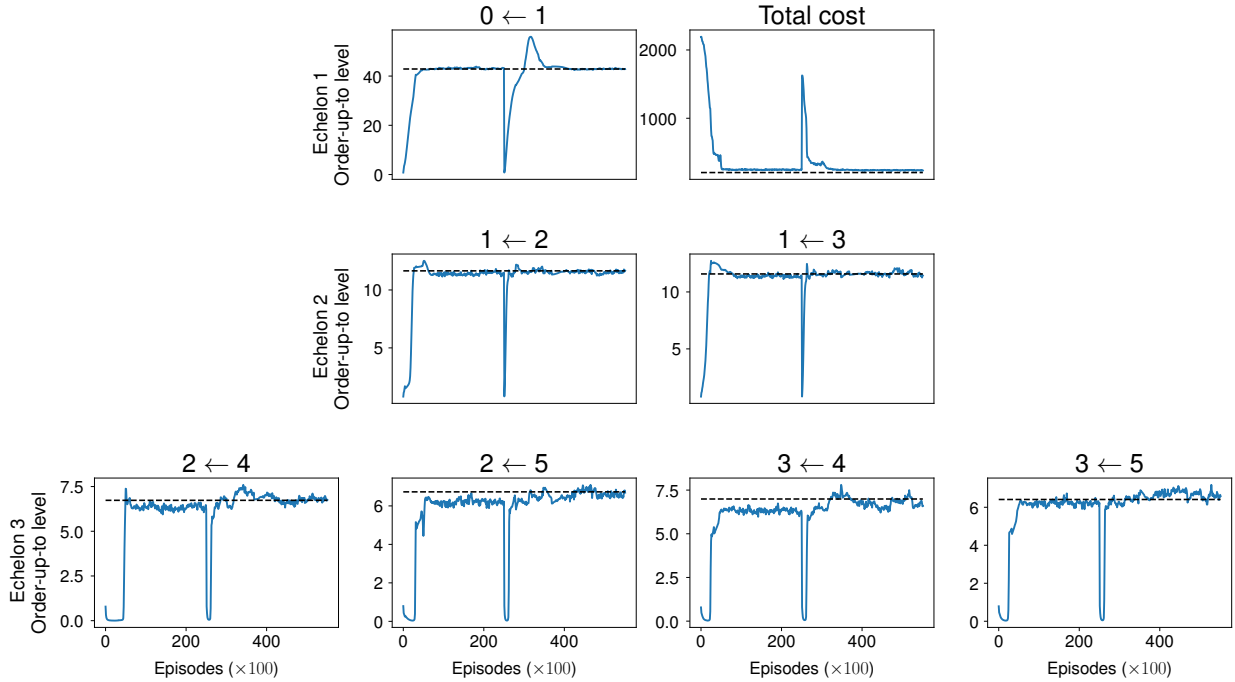}}
	\caption{Total cost and OUL decisions for mixed SMEIO.}
	\label{fig:mixedSCNResult}
\end{figure*}

From an inventory optimization perspective, there are no known analytical solutions for mixed SCNs, other than computationally intensive, enumeration-based approaches. Therefore, providing a stable numerical approach is greatly desirable. In this section, we demonstrate the performance of our method in finding OULs for multiple nodes of a mixed SCN simultaneously. We consider the SCN illustrated in Figure \ref{mixedSCN}. The network has 3 echelons, 5 nodes, 7 edges and two customers having independent stochastic normal distributions $\mathcal{N}(5,1)$ as their demands. The third echelon nodes, $4$ and $5$, are ``assembly-{\em and}'' nodes (see the definition in Section \ref{Model}.), and nodes 2 and 3 are distribution nodes, making the SCN a mixed one. We consider a time horizon of $T=10$. All 7 OULs are required to be optimized simultaneously. Table \ref{tab4} shows the parameters used in the study. Figure \ref{fig:mixedSCNResult} illustrates the convergence curves of the DNN approach. The black dotted lines are the best results, corresponding to the minimum costs achieved by DNN.

We allow the algorithm to restart the learning procedure from scratch using the best OULs found previously as the new initial inventory levels until there is no extra improvement in the objective value. We do this for two reasons. First, we do not know whether the objective functions of mixed SCN inventory problems are convex or nonconvex. In case we are minimizing a nonconvex objective, we do not wish the optimization algorithm to be stuck in a local minimum. Second, we have no idea how far away the initial inventory levels are from the optimal OULs. Considering that we are dealing with finite-horizon ($T=10$) episodes, this can significantly slow the method's learning procedure. Hence, we use the OULs found the previous time as the initial inventory levels of the next learning procedure. We set stopping criteria so that the procedure terminates when there is less than a 1\% change in the objective value.

One can observe the following based on Figure \ref{fig:mixedSCNResult}:
\begin{itemize}
	\item The jumps in the subplots are due to the restart procedure discussed previously. The importance of better initialization can be seen in the total cost spike at the restart. Although the OUL decisions are random and not necessarily close to their optimal values, at the restart, the total cost is almost half of the cost at the beginning of the procedure.  
	\item The farther downstream the node is, the noisier the OULs are.
	\item Similar to the previous examples, the agents farther upstream need fewer episodes to be optimized compared with those downstream.
\end{itemize}

\begin{table*}[]
	\caption{Mixed SCN parameters}
	\label{tab4}
	\setlength\tabcolsep{2pt}
	\centering	
	\begin{tabular}{ccccccccccccc}
		Echelon &  & Edge  &  & Holding cost &  & Shortage Cost & & Shipment lead-time &  & Order lead-time &  & Initialization \\
		&  &   &  & (per item) &  & (per item) & &     &  &  &  &  \\
		\cline{1-1} \cline{3-3} \cline{5-5} \cline{7-7} \cline{9-9} \cline{11-11} \cline{13-13}
		1       &  & (0,1) &  & $2$          &  & $4$          & & $2$                &  & $0$             &  & $40$           \\
		2       &  & (1,2) &  & $4$          &  & $12$         & & $1$                &  & $0$             &  & $10$           \\
		2       &  & (1,3) &  & $4$          &  & $12$         & & $1$                &  & $0$             &  & $10$           \\
		3       &  & (2,4) &  & $7$          &  & $37.12$      & & $1$                &  & $0$             &  & $5$            \\
		3       &  & (2,5) &  & $7$          &  & $37.12$      & & $1$                &  & $0$             &  & $5$            \\
		3       &  & (3,4) &  & $7$          &  & $37.12$      & & $1$                &  & $0$             &  & $5$            \\
		3       &  & (3,5) &  & $7$          &  & $37.12$      & & $1$                &  & $0$             &  & $5$            \\ \hline
	\end{tabular}
\end{table*}

Because there are no reliable algorithms for general mixed systems in the literature, we use a randomized search as a benchmark to compare with the DNN-SMEIO framework. We consider 100 separate randomly generated solutions and for each run we average over 2000 episodes. (Note that this is more than 20 times the number of environment interactions that the DNN approach requires; recall that DNN-SMEIO reaches its best results in at most $10^4$ episodes.)

We select candidate solutions randomly by setting each node's OUL equal to the mean demand of the node plus the absolute value of a random, zero-mean, normal random variate. We use random variates with larger standard deviations for nodes with larger demand means to allow for a search space that is likely to contain good solutions. Table \ref{tab5} shows the parameters for the random variates, as well as a comparison between the 5 best solutions found using this method and the solution found by the DNN approach. As one can see, the DNN performs better than all of the randomly generated solutions. 

\begin{table*}[]
	\caption{Top 5 randomized solutions and DNN results comparison for mixed SCN}
	\label{tab5}
	\setlength\tabcolsep{2pt}
	\centering	
	\begin{tabular}{cccccccccccccccccc}
		& & & & & & & \multicolumn{11}{c}{OULs} \\ 
		\cline{8-18}
		& Echelon &  & Edge  &  & Random search parameters      &  & \multicolumn{9}{c}{5 best randomly generated solutions} &  & DNN \\ \cline{2-2} \cline{4-4} \cline{6-6} \cline{8-16} \cline{18-18} 
		& 1       &  & (0,1) &  & $40 + | \mathcal{N}(0,4)|$ &  & 40.00   &   & 43.58   &   & 41.22  &  & 45.50  &  & 41.19  &                   & 42.87                \\
		& 2       &  & (1,2) &  & $10 + | \mathcal{N}(0,2)|$ &  & 13.54   &   & 12.84   &   & 10.57  &  & 12.49  &  & 13.07  &                   & 11.65                \\
		& 2       &  & (1,3) &  & $10 + | \mathcal{N}(0,2)|$ &  & 13.72   &   & 12.02   &   & 11.57  &  & 10.62  &  & 13.21  &                   & 11.58                \\
		& 3       &  & (2,4) &  & $5 +  | \mathcal{N}(0,2)|$ &  & 6.31    &   & 7.58    &   & 5.08   &  & 6.99   &  & 8.03   &                   & 6.73                 \\
		& 3       &  & (2,5) &  & $5 +  | \mathcal{N}(0,2)|$ &  & 6.50    &   & 5.85    &   & 5.77   &  & 5.36   &  & 9.07   &                   & 6.73                 \\
		& 3       &  & (3,4) &  & $5 +  | \mathcal{N}(0,2)|$ &  & 5.84    &   & 7.79    &   & 7.54   &  & 6.36   &  & 5.14   &                   & 6.99                 \\
		& 3       &  & (3,5) &  & $5 +  | \mathcal{N}(0,2)|$ &  & 5.16    &   & 5.16    &   & 9.45   &  & 5.39   &  & 8.00   &                   & 6.41                 \\
		\hline
		& Total Cost                         &  &       &  &                            &  & 215.05  &   & 214.72  &   & 214.45 &  & 212.97 &  & 211.90 &                   & 208.80              \\ 
	\end{tabular}
\end{table*}

We compare the performance of the proposed method with two additional alternatives: derivative-free optimization (DFO) \cite{conn2009introduction} and Spearmint Bayesian optimization \cite{swersky2013multi}. Each DFO or Spearmint step can be reduced to three parts: first, a suggestion of OULs; second, an independent simulation run to obtain the objective value for that suggestion; third, a suggestion of a new set of OULs based on an optimization algorithm. We consider 2000 episodes to allow each simulation run to converge. We test two cases, one in which there is no explicit bound on the number of function evaluations and one in which we restrict them to make sure the number of interactions between agents and the environment stays equal for the DFO, Spearmint, and DNN approaches. For this instance, 25 function evaluations for DFO or Spearmint were conducted, each of which consists of 2000 episodes, so we allowed DNN to use 50,000 episodes to reach its result. The only stopping criteria for the cases without an upper bound on the number of function evaluations are either having 100 algorithm steps without any improvement or having the last 10 improvements be less than 0.5\% of the cost function value. In addition, DFO-TR requires an initial guess to start the algorithm. We use the lead-time demand means for this purpose. Spearmint, however, requires an interval for each decision variable. We consider lead-time demand means as the lower bounds and an acceptable range (at least twice the lead-time demand standard deviation) to cover possible solutions. Both algorithms might suffer from a dependence on these initial conditions.

Table \ref{tab5-2} shows the comparison results. DFO and Spearmint find marginally better values in terms of objective function value for the cases without any upper bound on the number of function evaluations. However, restricting them to have the same number of interactions with the environment as the DNN has results in inferior performance of these alternatives compared to the DNN. Furthermore, because the optimal OULs for this mixed SCN structure happen to be close to the lead-time demand means, this biases the experiment in favor of DFO and Spearmint, which are given the lead-time demand means (or a small interval containing them) as initial values. In the next section, however, we investigate a more realistic case study and explore the comparison further. 

\begin{table*}[]
	\caption{Comparison results between DFO, Spearmint and DNN approaches for Mixed SCN}
	\label{tab5-2}
	\hspace*{-1.5cm}
	\setlength\tabcolsep{1.5pt}
	\centering	
	\begin{tabular}{cccccccccccccccccc}
		&         &  &       &  & \multicolumn{11}{c}{Alternatives}                                                                                                                                   &  & \multirow{3}{*}{DNN} \\ \cline{6-16}
		&         &  &       &  & \multicolumn{3}{c}{Initial Value Choosing} &  & \multicolumn{3}{c}{Results with 25 function evaluations} &  & \multicolumn{3}{c}{Best results without restrictions} &  &                      \\ \cline{6-8} \cline{10-12} \cline{14-16}
		\multirow{8}{*}{OUL decisions} & Echelon &  & Edge  &  & DFO          &          & Spearmint        &  & DFO                &              & Spearmint            &  & DFO               &             & Spearmint           &  &                      \\ \cline{2-2} \cline{4-4} \cline{6-6} \cline{8-8} \cline{10-10} \cline{12-12} \cline{14-14} \cline{16-16}
		& 1       &  & (0,1) &  & $40$         &          & $[40,48]$        &  & 47.69              &              & 41.45                &  & 43.73             &             & 43.80               &  & 42.87                 \\
		& 2       &  & (1,2) &  & $10$         &          & $[10,14]$        &  & 12.45              &              & 12.34                &  & 11.46             &             & 11.45               &  & 11.65                 \\
		& 2       &  & (1,3) &  & $10$         &          & $[10,14]$        &  & 12.62              &              & 11.76                &  & 11.46             &             & 11.49               &  & 11.58                 \\
		& 3       &  & (2,4) &  & $5$          &          & $[5,9]$          &  & 5.51               &              & 5.39                 &  & 5.77              &             & 5.80                &  & 6.73                  \\
		& 3       &  & (2,5) &  & $5$          &          & $[5,9]$          &  & 5.58               &              & 5.54                 &  & 5.77              &             & 5.77                &  & 6.73                  \\
		& 3       &  & (3,4) &  & $5$          &          & $[5,9]$          &  & 5.53               &              & 7.02                 &  & 5.77              &             & 5.78                &  & 6.99                  \\
		& 3       &  & (3,5) &  & $5$          &          & $[5,9]$          &  & 5.40               &              & 5.63                 &  & 5.78              &             & 5.78                &  & 6.41                  \\
		Total Cost                             &         &  &       &  &              &          &                  &  & 215.21             &              & 214.66               &  & 206.35            &             & 206.36              &  & 208.80               \\ \hline
	\end{tabular}
\end{table*}

\subsection{Complex SCN}

\begin{figure*}
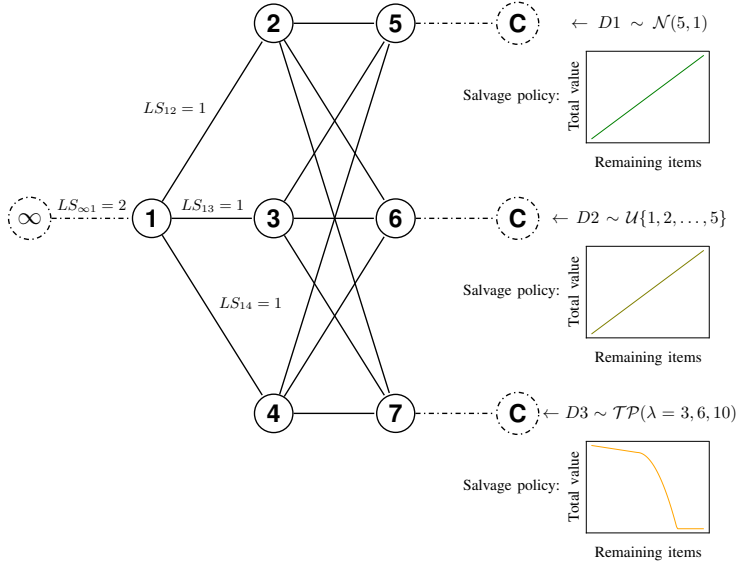

	\centering
\begin{tikzpicture}[shorten >=1pt, auto, node distance=3cm, thick,
   node_style/.style={circle,draw=black,font=\sffamily\Large\bfseries},
   node_style2/.style={circle,dashdotted,draw=black,font=\sffamily\Large\bfseries},
   edge_style/.style={draw=black,  thick},edge_style2/.style={draw=black,dashdotted,  thick},edge_style3/.style={draw=black,densely dashed,  thick}
   ]
    \tikzstyle{annot} = [text width=12em, text centered]

    \node[node_style2] (v0) at (-8.5,0) {$\infty$};
    \node[node_style] (v1) at (-6,0) {1};
    \node[node_style] (v2) at (-3.5,4) {2};
    \node[node_style] (v3) at (-3.5,0) {3};
    \node[node_style] (v4) at (-3.5,-4) {4};
    \node[node_style] (v5) at (-1,4) {5};
    \node[node_style] (v6) at (-1,0) {6};        
    \node[node_style] (v7) at (-1,-4) {7};
    \node[node_style2] (vc1) at (1.5,4) {C};
	\node[node_style2] (vc2) at (1.5,0) {C};
	\node[node_style2] (vc3) at (1.5,-4) {C};
    \draw[edge_style2]  (v0) edge node{\small $LS_{\infty1}=2$} (v1);
    \draw[edge_style2]  (v5) edge node{} (vc1);
	\draw[edge_style2]  (v6) edge node{} (vc2);
    \draw[edge_style2]  (v7) edge node{} (vc3);
    \draw[edge_style]   (v1) edge node{\small$LS_{12}=1$} (v2);
    \draw[edge_style]   (v1) edge node{\small$LS_{13}=1$} (v3);
    \draw[edge_style]   (v1) edge node{\small$LS_{14}=1$} (v4);
    \draw[edge_style]   (v2) edge node{} (v5);
    \draw[edge_style]   (v2) edge node{} (v6);
    \draw[edge_style]   (v2) edge node{} (v7);
    \draw[edge_style]   (v3) edge node{} (v5);
	\draw[edge_style]   (v3) edge node{} (v6);
	\draw[edge_style]   (v3) edge node{} (v7);
    \draw[edge_style]   (v4) edge node{} (v5);
	\draw[edge_style]   (v4) edge node{} (v6);
	\draw[edge_style]   (v4) edge node{} (v7);
	
    \node[annot,right of=vc1, node distance=2.5cm] (demand1) {\bf $\leftarrow D1 \sim \mathcal{N}(5,1)$ }; 
    \node[annot,below of=demand1, node distance=2.87cm] (item1) {\footnotesize   \ \ \    Remaining items };

    \node[annot,right of=vc2, node distance=2.5cm] (demand2) {\bf $\leftarrow D2 \sim \mathcal{U}\{1,2,\dots,5\}$ }; 
        \node[annot,below of=demand2, node distance=2.87cm] (item1) {\footnotesize   \ \ \    Remaining items };
    \node[annot,right of=vc3, node distance=2.5cm] (demand3) {\bf $\leftarrow D3 \sim \mathcal{TP}(\lambda =3,6,10)$ }; 
    \node[annot,below of=demand3, node distance=2.87cm] (item1) {\footnotesize   \ \ \    Remaining items };

	\node [matrix, node distance=1.7cm]  at (2.75,2.5) (Salvage1) { \node [annot] at (5.25,1) (policy) {  \small Salvage policy:};  \node [annot] at (8,1) (policy) { \scalebox{0.2}{\input{sal1.pgf}}  };\\ }; 
    \node[annot,right of=Salvage1, rotate=90, node distance=-0.1cm] (item1) {\footnotesize       Total value };	
	\node [matrix, node distance=1.7cm] at (2.75,-1.5) (Salvage2) { \node [annot] at (5.25,1) (policy) {  \small Salvage policy:};  \node [annot] at (8,1) (policy) { \scalebox{0.2}{\input{sal2.pgf}}  };\\ }; 
    \node[annot,right of=Salvage2, rotate=90, node distance=-0.1cm] (item1) {\footnotesize     Total value };	
	\node [matrix, node distance=1.7cm] at (2.75,-5.5) (Salvage3) { \node [annot] at (5.25,1) (policy) {  \small Salvage policy:};  \node [annot] at (8,1) (policy) { \scalebox{0.2}{\input{sal3.pgf}}  };\\ }; 		
    \node[annot,right of=Salvage3, rotate=90, node distance=-0.1cm] (item1) {\footnotesize      Total value };	   
    	

    \end{tikzpicture}
	\caption{Mixed SCN with customer demands and salvage policies.}
	\label{fig:caseStudy}
\end{figure*}

In this section we introduce a case study of a general-structured SCN with realistic settings such as nonlinear holding and penalty costs and salvage values. For this comparison, We consider trust-region derivative-free optimization (DFO-TR) \cite{bandeira2012computation,pirhooshyaran2020feature}, GPEI Bayesian along with simple yet powerful random search techniques as alternative methods. We consider the complex SCN shown in Figure \ref{fig:caseStudy}. The SCN contains 7 nodes and 13 edges. Nodes 5, 6 and 7 are ``assembly-{\em and}'' nodes. That is, to produce one item at any of these nodes, one item from each of its predecessors is required. In this case study, holding costs are considered to be piecewise linear instead of linear. When the number of items in inventory increases beyond a certain threshold, the cost per item decreases. On the other hand, shortage penalty costs are considered to be nonlinear, and the cost per item increases when the shortage quantity is greater. There are three leaf nodes that see customer demands. One follows an independent normal distribution $\mathcal{N}(5,1)$, the second follows a discrete uniform distribution $\mathcal{U}\{1,2,\dots,5\}$, and the third follows a two-sided truncated Poisson distribution $\mathcal{TP}(\lambda =3;6;10)$, where 3 is the distribution parameter and 6 and 10 are the beginning and the end of the possible values, respectively. We consider $T=10$ as the episode horizon length. 

We assume that the three leaf nodes have different salvage functions that represent the cost or reward incurred based on the inventory level at the end of the horizon. Nodes 5 and 6 have linear rewards, with different slopes; for example, these might model situations in which excess inventory can be sold for a per-unit cost at the end of the horizon. However, for node 7, we assume a nonlinear salvage reward policy to clear the remaining inventory. Up to a threshold, the price per item is considered to be high because there are only few items left. The price per item then reduces drastically to reach the second threshold and thereafter it is fixed. 

Table \ref{tab6} provides further details about this problem instance. In the table, $(x<\text{threshold},f_1(x),f_2(x))$ means that when $x < \text{threshold}$, the function $f_1(x)$ is used, and afterwards $f_2(x)$ is used. The initial inventories are set equal to the demand mean. For instance, $\mathbb{E}\left( \mathcal{TP}(\lambda =3,6,10) \right) = 7.58$ is considered as the initialization values for the $(2\leftarrow7),(3\leftarrow7)$ and $(4\leftarrow7)$ decisions.

\begin{table*}[]
	\caption{Case study mixed SCN parameters$^\ast$}
	\label{tab6}
	\setlength\tabcolsep{2pt}
	\hspace*{0.25cm}
	\begin{small}
		\begin{tabular}{ccccccccccc}
			Edge  &  & Holding cost &  & Shortage Cost & & Shipment lead-time &  & Salvage reward &  & Initialization \\
			&  & (for $x$ units on hand) &  & (for $x$ backorders) & &     &  & (for $x$ units on hand) &  &  \\
			\cline{1-1} \cline{3-3} \cline{5-5} \cline{7-7} \cline{9-9} \cline{11-11} 
			(0,1) &  & $2x$                   &  & $4x$                     & & $2$                &  & ---                                        &  & $45.24$           \\
			(1,2) &  & $(x<3,4x,3x)$          &  & $(x<3,12x,4x^2)$         & & $1$                &  & ---                                             &  & $15.08$           \\
			(1,3) &  & $(x<3,4x,3x)$          &  & $(x<3,12x,4x^2)$         & & $1$                &  & ---                                            &  & $15.08$           \\
			(1,4) &  & $(x<3,4x,3x)$          &  & $(x<3,12x,4x^2)$         & & $1$                &  & ---                                           &  & $15.08$           \\		
			(2,5) &  & $(x<3,7x,6x)$          &  & $(x<3,36x,12x^2)$      & & $1$                &  & $1.25x$                                           &  & $5.00$            \\
			(2,6) &  & $(x<3,7x,6x)$          &  & $(x<3,36x,12x^2)$      & & $1$                &  & $1.5x$                                           &  & $2.50$            \\
			(2,7) &  & $(x<3,7x,6x)$          &  & $(x<3,36x,12x^2)$      & & $1$                &  & $(x<2,15-0.5x,max(-3.5x^2 +14x,3))$             &  & $7.58$            \\
			(3,5) &  & $(x<3,7x,6x)$          &  & $(x<3,36x,12x^2)$      & & $1$                &  & $1.25x$                                        &  & $5.00$            \\
			(3,6) &  & $(x<3,7x,6x)$          &  & $(x<3,36x,12x^2)$      & & $1$                &  & $1.5x$                                        &  & $2.50$            \\		
			(3,7) &  & $(x<3,7x,6x)$          &  & $(x<3,36x,12x^2)$      & & $1$                &  & $(x<2,15-0.5x,max(-3.5x^2 +14x,3))$             &  & $7.58$            \\ 
			(4,5) &  & $(x<3,7x,6x)$          &  & $(x<3,36x,12x^2)$      & & $1$                &  & $1.25x$                                            &  & $5.00$            \\
			(4,6) &  & $(x<3,7x,6x)$          &  & $(x<3,36x,12x^2)$      & & $1$                &  & $1.5x$                                           &  & $2.50$            \\		
			(4,7) &  & $(x<3,7x,6x)$          &  & $(x<3,36x,12x^2)$      & & $1$                &  & $(x<2,15-0.5x,max(-3.5x^2 +14x,3))$             &  & $7.58$            \\ \hline
		\end{tabular}
		$^\ast$ {\footnotesize All of the costs listed below equal 0 if $x < 0$.}
	\end{small}
\end{table*}

First, we compare the DNN-SMEIO method with the randomized approach. We consider 400 runs of 5000 episodes each. Table \ref{tab7} shows the results of this comparison. The intervals of possible OULs used for the randomized approach contain the best solution found by DNN-SMEIO. One can see the clear advantage of the proposed method over the randomized approach. 

\begin{table*}[]
	\caption{Top 5 randomized solutions and DNN results comparison for mixed SCN}
	\label{tab7}
	\setlength\tabcolsep{2pt}
	\centering	
	\begin{small}
		\begin{tabular}{cccccccccccccccccc}
			& & & & & & & \multicolumn{11}{c}{OULs} \\
			\cline{8-18}
			& Echelon &  & Edge  &  & Random search parameters      &  & \multicolumn{9}{c}{5 best randomly generated solutions} &  & DNN \\ \cline{2-2} \cline{4-4} \cline{6-6} \cline{8-16} \cline{18-18}
			
			& 1       &  & (0,1) &  & $45.24 + |\mathcal{N}(0,50)|$ &  & 102.73  &   & 102.56  &   & 111.44 &  & 100.40 &  & 100.40  &                    & 101.44                \\
			& 2       &  & (1,2) &  & $15.08 + | \mathcal{N}(0,5)|$ &  & 19.50   &   & 17.96   &   & 22.71   &  & 22.75  &  & 15.91  &                    & 18.75                \\
			& 2       &  & (1,3) &  & $15.08 + | \mathcal{N}(0,5)|$ &  & 14.25   &   & 14.51   &   & 18.72   &  & 18.81  &  & 19.84  &                    & 20.82                \\
			& 2       &  & (1,4) &  & $15.08 + | \mathcal{N}(0,5)|$ &  & 15.51   &   & 15.10   &   & 20.05   &  & 20.51  &  & 21.85  &                    & 21.48                \\
			& 3       &  & (2,5) &  & $5 +     | \mathcal{N}(0,5)|$ &  & 7.52    &   & 6.91    &   & 6.06    &  & 5.33   &  & 6.26   &                    & 7.29                 \\
			& 3       &  & (2,6) &  & $2.5 +   | \mathcal{N}(0,5)|$ &  & 7.75    &   & 10.38   &   & 5.51    &  & 8.95   &  & 7.16   &                    & 6.95                 \\
			& 3       &  & (2,7) &  & $7.58 +  | \mathcal{N}(0,5)|$ &  & 9.38    &   & 8.83    &   & 12.52   &  & 12.29  &  & 9.84   &                    & 10.28                 \\
			& 3       &  & (3,5) &  & $5 +     | \mathcal{N}(0,5)|$ &  & 8.03    &   & 7.87    &   & 9.65    &  & 9.39   &  & 6.47   &                    & 7.21                 \\
			& 3       &  & (3,6) &  & $2.5 +   | \mathcal{N}(0,5)|$ &  & 6.99    &   & 6.55    &   & 4.08    &  & 5.19   &  & 5.73   &                    & 6.06                 \\
			& 3       &  & (3,7) &  & $7.58 +  | \mathcal{N}(0,5)|$ &  & 11.87   &   & 11.14   &   & 8.17    &  & 8.52   &  & 9.17   &                    & 9.48                 \\
			& 3       &  & (4,5) &  & $5 +     | \mathcal{N}(0,5)|$ &  & 7.04    &   & 7.23    &   & 9.11    &  & 5.16   &  & 5.02   &                    & 6.11                 \\
			& 3       &  & (4,6) &  & $2.5 +   | \mathcal{N}(0,5)|$ &  & 8.51   &   & 6.70    &   & 8.85    &  & 5.54   &  & 5.15   &                    & 6.5                 \\
			& 3       &  & (4,7) &  & $7.58 +  | \mathcal{N}(0,5)|$ &  & 10.10   &   & 9.84    &   & 10.48   &  & 8.25   &  & 7.71   &                    & 9.38                 \\ \hline
			& Total Cost                         &  &       &  &    &  & 534.06  &   & 526.75  &   & 522.18  &  & 519.68 &  & 514.69 &                    & 478.61               \\
			
		\end{tabular}
	\end{small}
\end{table*}

We further investigate a comparison between DFO, Spearmint, and the proposed DNN approach. We do not put an upper limit on the number of interactions with the environment for the DFO and Spearmint approaches. (This biases the experiment in favor of alternative approaches.) Table \ref{tab8} tabulates the results. One can see the clear advantage of the proposed method over the alternatives. Both DFO and Spearmint fail to find results that are nearly as good as the randomized method. We also emphasize the fact that choosing a suitable interval for the Spearmint method requires expert knowledge, trial and error or multiple runs of the algorithm. For instance, based on the results reported in Table \ref{tab7}, we already knew that OULs greater than 100 should be considered for the $0\leftarrow 1$ decision. Consequently, we assumed a large interval of $[45, 150]$ for this decision variable, but without this prior knowledge, one might miss important regions for the decision variable. We refer the reader to see the learning curves and loss behavior plus their discussions in Section \textcolor{blue}{5} of supplement.     

\begin{table*}[]
	\caption{Comparison results between DFO, Spearmint and DNN approaches for case study SCN}
	\label{tab8}
	\setlength\tabcolsep{2pt}
	\centering	
	\begin{small}
		\begin{tabular}{cccccccccccccc}
			&         &  &       &  & \multicolumn{7}{c}{Alternatives}                                            &  & \multirow{3}{*}{DNN} \\ \cline{6-12}
			&         &  &       &  & \multicolumn{3}{c}{Initializing} &  & \multicolumn{3}{c}{results} &  &                      \\ \cline{6-8} \cline{10-12}
			\multirow{8}{*}{OUL decisions} & Echelon &  & Edge  &  & DFO          &          & Spearmint        &  & DFO      &    & Spearmint   &  &                      \\ \cline{1-1}\cline{2-2} \cline{4-4} \cline{6-6} \cline{8-8} \cline{10-10} \cline{12-12}
			& 1       &  & (0,1) &  & $45.24$     & & $[45, 150]$ & & 99.15 & & 93.86 & & 101.44                \\
			& 2       &  & (1,2) &  & $15.08$     & & $[15, 45]$ & & 15.94 & & 29.93  & & 18.75                 \\
			& 2       &  & (1,3) &  & $15.08$     & & $[15, 45]$ & & 16.01 & & 24.40  & & 20.82                 \\
			& 2       &  & (1,4) &  & $15.08$     & & $[15, 45]$ & & 15.99 & & 26.01    & & 21.48                 \\
			& 3       &  & (2,5) &  & $5$         & & $[2.5, 15]$ & & 5.20 & & 5.08   & & 7.29                  \\
			& 3       &  & (2,6) &  & $2.5$       & & $[2.5, 15]$ & & 3.13 & & 3.24   & & 6.95                  \\
			& 3       &  & (2,7) &  & $7.58$      & & $[2.5, 15]$ & & 7.62 & & 8.02   & & 10.28                 \\
			& 3       &  & (3,5) &  & $5$         & & $[2.5, 15]$ & & 5.41 & & 5.21   & & 7.21                  \\
			& 3       &  & (3,6) &  & $2.5$       & & $[2.5, 15]$ & & 2.84 & & 3.41   & & 6.06                  \\
			& 3       &  & (3,7) &  & $7.58$      & & $[2.5, 15]$ & & 7.52 & & 8.22   & & 9.48                  \\
			& 3       &  & (4,5) &  & $5$         & & $[2.5, 15]$ & & 5.44 & & 5.17   & & 6.11                  \\
			& 3       &  & (4,6) &  & $2.5$       & & $[2.5, 15]$ & & 2.77 & & 3.37   & & 6.5                   \\
			& 3       &  & (4,7) &  & $7.58$      & & $[2.5, 15]$ & & 7.52 & & 8.37   & & 9.38                  \\
			Total Cost                             &         &  &       &  &              &          &                  &  & 644.41   &    & 618.44      &  & 478.61               \\ \hline
		\end{tabular}
	\end{small}
\end{table*}

\begin{table}[]
	\caption{Cost comparisons for complex mixed SCN instances}
	\label{TabComplex}
	\setlength\tabcolsep{3pt}
	\centering	
	\begin{small}
		\begin{tabular}{ccc}
			Instance & DNN-SMIO  & DFO \\  \hline
			1 & 380.95 & 402.41 \\ 
			2 & 419.13 & 442.42 \\
			3 & 407.83 & 408.27 \\
			4 & 379.31 & 400.04 \\
			5 & 478.61 & 644.41 \\ \hline
			average & 426.43 & 446.32   \\
		\end{tabular}
	\end{small}
\end{table}

We further conduct a cost comparison between the DFO and DNN approaches for 4 new instances based on the complex SCN structure shown in Figure \ref{fig:caseStudy}. Table \ref{TabComplex} shows the cost comparison results for the new instances plus the one already discussed earlier on the structure. We focused on different salvage costs and demand distributions for this comparison. We refer the reader to Table \textcolor{blue}{5} of the supplement for the structure details and suggested OULs. As can be seen, the DNN approach outperforms DFO for all five cases.

\section{Conclusion} 
\label{Conclusion}

This research studies simultaneous decision-making for stochastic multi-echelon inventory optimization with arbitrary SCN topologies, demand distributions, and cost structures, using deep neural networks as decision makers, considering a finite-horizon. We introduce pairwise modeling of SMEIOs and associate a DNN to each edge in need of decision making. The DNNs constantly interact with their environment (the supply chain network) and aim to learn the OULs minimizing the total network cost. We assume that the demand distribution as well as the inventory levels are known to all agents. Our research is one of the first works considering deep neural networks as joint decision makers in an SMEIO framework that can suggest clear and interpretable OULs as an output.

The findings indicate the effectiveness of the method both in terms of its accuracy compared to analytical exact solutions and versus alternatives such as DFO and Spearmint Bayesian optimization, as well as  in terms of the computational expense (interactions with environment) compared to enumeration methods. The convergence curves shown for the single-node newsvendor, serial, and mixed systems validate the stability of the framework. For mixed supply networks with more advanced cost schemes, findings suggest that upstream echelons hold more items compared to their demand means than downstream echelons. For example, the OUL found by DNN-SMEIO for the first echelon $(0\leftarrow 1)$ in the complex SCN shown in Figure \ref{fig:caseStudy} is more than twice the demand it sees.

This study can be extended to cases in which the demands are auto-correlated and/or the decision makers/DNNs have partial information about the structure of the SCN. Another future study can be conducted to investigate closed-loop supply chain networks. In addition, considering time dependent OULs would be a desirable improvement to this method.

	\bibliography{RLSCPaper}

	\bibliographystyle{unsrtnat}

\end{document}


\maketitle
	
	\section{Stochastic multi-echelon inventory optimization}
	
	Research on SMEIO can be classified according to several dimensions. The number of echelon layers as well as their topological structure are essential factors. One can categorize SMEIO into single-, two-, and multi-echelon networks with serial, divergent, convergent, or general structures \cite{fosct2e}. Furthermore, inventory review can be continuous or periodic, meaning that the system's inventory levels evolve smoothly over time or update once in each decision period. Demands can be discrete (mainly Poisson \cite{topan2017heuristics}, compound Poisson \cite{johansson2020controlling}, and Erlang \cite{bitton2019joint}), continuous (mainly normal \cite{cobb2016lead}) or general stochastic \cite{sapra2017dual}.
	
	Solution approach is another dimension along which to classify SMEIO. Simulation--optimization \cite{xu2019simulation}, computational and numerical \cite{oroojlooyjadid2017deep}, approximate \cite{topan2017heuristics}, and exact (theoretical) methods \cite{sapra2017dual} are among the most common solution approaches. In most approaches, we impose a particular form of the inventory policy (e.g., a base-stock policy), and an algorithm is used to obtain a near-optimal set of parameters for that policy to optimize the objective function \cite{de2018typology}. This approach is especially prevalent for systems for which the optimal policy structure is unknown, a category which includes distribution systems \cite{federgruen1984approximations} as well as general networks. However, this approach has some limitations. The methods are dependent on the problem settings, and adjustments in the chosen policy or system structure require the algorithm to be initiated once again. We aim to reduce these limitations by using trainable DNN algorithms in which one can transfer the knowledge (learned weights of agents' networks) from a SCN to a modified one.
	
	Most SMEIO models fall either into the category of {\em stochastic service models} (SSM) or of {\em guaranteed service models} (GSM). SSM models allow for the possibility of stockouts at various nodes of the SMEIO system, which then result in stochastic lead times to their downstream nodes. All of the models we have discussed so far are SSM models, as is the approach we propose. In contrast, GSM models impose an upper bound on the demand, which allows all nodes in the system to guarantee delivery within a fixed number of time periods. This guarantee leads to greater tractability (and therefore greater adoption in practice), even for general network topologies, but also requires some strong assumptions, the demand bound chief among them. To the best of our knowledge, no GSM models in the literature can accommodate general holding and stockout cost structures, as our approach can. Finally, we note that the inventory dynamics in SSM and GSM models operate essentially in the same way; the difference lies in how the base-stock levels are optimized, rather than in how the systems are managed. For further discussion of SSM and GSM models, see \cite{GrWi03chapter,fosct2e}.
	\section{Summary of Notation}
	The notation is summarized in Table~\ref{table:notation} of Supplementary.
	
	\section{Framework Structure, Optimization Algorithm and Hyperparameter Tuning} 
	
	\begin{table}
		\caption{Notation summary.}
		\label{table:notation}
		\centerline{
			\begin{tabular}{lll}
				\hline
				\multicolumn{3}{l}{Inputs} \\
				& $T$ & Length of the planning horizon \\
				& $\mathcal{N}, \mathcal{E}$ & Sets of nodes and edges in SCN \\
				& $\mathcal{U}_j$, $\mathcal{D}_j$ & Sets of nodes upstream from and downstream from (respectively) node $j$ \\
				& $h_{ij}(\cdot)$ & Holding cost function for items from node $i$ in raw material, finished goods, or in-transit inventory at node $j$ \\
				& $p_{ij}(\cdot)$ & Stockout cost function for stockouts at node $i$ that are owed to node $j$ \\
				& $L_{jk}$ & Lead time for items shipped from node $j$ to node $k$ \\
				\multicolumn{3}{l}{State Variables} \\
				& $D_{jkt}$ & Demand placed by node $k$ to node $j$ in period $t$ \\
				& $S_{jkt}$ & Number of units shipped from node $j$ to node $k$ in period $t$ \\
				& $IL_{jt}$ & Finished-goods inventory level at node $j$ at the end of period $t$ \\
				& $IL^r_{jit}$ & Raw-material inventory level at node $j$ for items from node $i$ at the end of period $t$ \\
				& $BO_{jkt}$ & Backorders at node $j$ owed to node $k$ at the end of period $t$ \\
				& $IT_{jkt}$ & Inventory in transit from node $j$ to node $k$ at the end of period $t$ \\
				& $IP_{jit}$ & Inventory position of item-$i$ materials at node $j$ in period $t$ immediately before order is placed \\
				\multicolumn{3}{l}{Decision Variables} \\
				& $OUL_{ji}$ & Order-up-to level (base-stock level) used by node $j$ when placing orders from node $i$ \\
				\hline
			\end{tabular}
		}
	\end{table}
	
	\begin{figure}[]
		\centering
			
\tikzstyle{decision} = [diamond,   draw, fill=white!20, text width=6em,  text centered,  node distance=3cm, inner sep=0pt]
\tikzstyle{block0}   = [rectangle, draw, fill=white!20, text width=25em, text centered,  node distance=3cm, rounded corners, minimum height=4em]
\tikzstyle{block00}  = [rectangle, draw, fill=white!20, text width=50em, text centered,  node distance=3cm, rounded corners, minimum height=4em]
\tikzstyle{block}    = [rectangle, draw, fill=white!20, text width=9em,  text centered,                     rounded corners, minimum height=4em]
\tikzstyle{block1}   = [rectangle, draw, fill=white!20, text width=16em, text centered,                     rounded corners, minimum height=4em]
\tikzstyle{line}     = [draw, -latex']
\tikzstyle{invisibleblock}    = [rectangle, fill=white!20, text width=10em,  text centered,  node distance=3cm,    rounded corners, minimum height=4em]

  \centering

  \begin{tikzpicture}[node distance = 2cm, auto]
    \node [block0, align=center] (init) { \begin{itemize} \item Initialize parameters \item Define $\mathcal{C}_k$ as the tuning loss function  \end{itemize}};
    \node [block00, align=center,below of = init] (GP) { {\bf Gaussian Process (GP)}: \\ Consider prior Gaussian distribution $y\sim \mathcal{N}(f(x),\nu)$ for hyperparameters where $\nu$ is the noise variance};
    \node [block, below of = GP, node distance = 2cm] (yes block1) {Train DNN-SMEIO environment \\({\bf loss}: $\mathcal{C}_k$)};
    \node [block, below of = yes block1, node distance = 2cm] (yes block2) {Find the up-to-order levels };
    \node [block, below of = yes block2, node distance = 2cm] (yes block3) {Evaluate the loss function};
    \node [block1, below of = yes block3, node distance = 2cm] (yes block4) {{\bf Expected Improvement (EI)}:\\ Find the next best feasible point};
    \node [block, left of = yes block1, node distance = 5cm] (Tr) {DNN-SMEIO train episodes};
	\node [block, left of = yes block2, node distance = 5cm] (Val) {DNN-SMEIO validation episodes };
	\node [invisibleblock, right of = yes block4, node distance =180](invis){While function evaluation budget left};
    \path [line] (init) -- (GP);
    \path [line] (GP) -- node {}(yes block1);
    \path [line] (yes block1) -- node {}(yes block2);
    \path [line] (yes block2) -- node {}(yes block3);    
    \path [line] (yes block3) -- node {}(yes block4);    
    \path [line] (yes block4) -- node {}(invis);
     \path [line] (invis) |- node {}(yes block1);           
    \path [line] (Tr) -- node {}(yes block1);
    \path [line] (Val) -- node {}(yes block2);

  \end{tikzpicture}
		\caption{Tuning DNN-SMEIO via GPEI Bayesian optimization}
		\label{Hyper}
	\end{figure}

	In this study, we utilize Spearmint Bayesian optimization, and in particular, the Gaussian process expected improvement (GPEI) method \cite{swersky2013multi}. Figure \ref{Hyper} shows a sketch of GPEI Spearmint. The number of shared layers (if any) between separate networks, the number of hidden layers for each network, the learning rate, the activation function type, the number of nodes in each layer and the size of the mini-batches are considered as the tunable hyperparameters.

	Table \ref{tab0} shows the values we considered for the parameters to be tuned. In most of our experiments, the best settings are to use totally separated DNNs (Multi-agent structure = 0) with Softplus activation functions, learning rate 0.01, and 4 hidden layers. The results appear to be quite quite insensitive to the mini-batch size.
	
	\begin{table}
		\hspace*{-1cm}
		\caption{Hyperparameter values for DNN-SMEIO framework.}
		\label{tab0}
		\vspace{1em}
		\begin{tabular}{clll}
			&	parameter & possible values & notes \\
			\hline
			1 & multi-agent structure & \{0,1\}& 0: totally separated, 1: sharing first layers                         \\ 
			2 &	\# hidden layers        & \{1,2,3,4,5\}  & total number of layers is this number plus 2                  \\
			3&	\# shared layers     & \{1,2,3\}      & only used if multi-agent structure = 1    \\
			4&	activation function       &  softplus, leakyReLU, ReLU &\\ 
			5&	\# nodes in each layer  & \{8,12,16,24,36,48\} &              \\ 
			6&	mini-batch size  & \{1,2,5,10\} & \\ 
			7& learning rate         &  \{0.0005,0.001,0.005,0.01,0.02\} &                            
		\end{tabular}
	\end{table}

	Because the episodes have a finite horizon, the initial values of the state variables can have a significant impact on the training process. Of course, in practice, one would set the initial values to the actual values of the inventory levels at the start of the horizon. For training, however, in each episode we chose to initialize the raw-material inventories to 0 and the finished-goods inventory levels to the lead-time demand. That is, we set $IL_{j0} = \mu_jL_j$ and $IL^r_{ji0}  = 0$ for all $j\in \mathcal{N}$ and $i\in \mathcal{U}_j$. The system is therefore ``primed'' with enough inventory to meet the upcoming expected demand.
	
	\section{Details of Assembly SCN Experiments}
	The complete specifications of the parameters of the assembly system instances in Section \textcolor{blue}{4.3} of the main article, as well as the OULs suggested by each method, are presented in Tables \ref{tabassem1} and \ref{tabassem2} of the supplement.

	\begin{table*}[]
		\caption{Assembly(1) SCN cases details}
		\label{tabassem1}
		\setlength\tabcolsep{4pt}
		\centering
		\begin{scriptsize}
			
			\begin{tabular}{llllllllll}
				Instance   1 &             &                    &               &                    &                                         &               &                    &               &               \\
				&             &                    &               &                    &                                         & \multicolumn{4}{c}{OUL}  \\
				Node         & Predecessor & H & S & LT          & Demand                                  & DNNSMIO       & CD & Enumeration   & DFO           \\
				1            & --          & 0.25               & 0             & 2                  & --                                      & 26.91         & 25.77            & 26.72       & 6.08        \\
				2            & --          & 0.25               & 0             & 2                  & --                                      & 26.8          & 25.77            & 26.72       & 27.32      \\
				3            & --          & 0.25               & 0             & 2                  & --                                      & 26.86         & 25.77            & 26.72       & 21.77      \\
				4            & --          & 0.25               & 0             & 2                  & --                                      & 26.85         & 25.77            & 26.72       & 26.89      \\
				5            & 1           & 0.8                & 0             & 1                  & --                                      & 13.55         & 13.87            & 13.36       & 12.20      \\
				5            & 2           & 0.8                & 0             & 1                  & --                                      & 13.57         & 13.87            & 13.36       & 14.75      \\
				6            & 3           & 0.8                & 0             & 1                  & --                                      & 13.58         & 13.87            & 13.36       & 12.85      \\
				6            & 4           & 0.8                & 0             & 1                  & --                                      & 13.57         & 13.87            & 13.36       & 15.36      \\
				7            & 5           & 1.9                & 10            & 1                  & N(13,(1.2)\textasciicircum{}2)          & 14.64         & 15.08            & 15.16       & 14.41      \\
				7            & 6           & 1.9                & 10            & 1                  & N(13,(1.2)\textasciicircum{}2)          & 14.64         & 15.08            & 15.16       & 14.46      \\
				&             &                    &               & \multicolumn{2}{l}{Cost} & 40.55       & 40.26            & 40.34       & 233.45        \\
				&             &                    &               &                    &                                         &               &                    &               &               \\
				Instance 2   &             &                    &               &                    &                                         &               &                    &               &               \\
				&             &                    &               &                    &                                         &               &                    &               &               \\
				&             &                    &               &                    &                                         &               &                    &               &               \\
				&             &                    &               &                    &                                         & \multicolumn{4}{c}{OUL}  \\
				Node         & Predecessor & H & S & LT          & Demand                                  & DNNSMIO       & CD & Enumeration   & DFO           \\
				1            & --          & 2                  & 0             & 2                  & --                                      & 10.08         & 7.50            & 7.5           & 10.19      \\
				2            & --          & 2                  & 0             & 2                  & --                                      & 10.08         & 7.50            & 7.5           & 11.83      \\
				3            & --          & 2                  & 0             & 2                  & --                                      & 10.13         & 7.50            & 7.5           & 1.00       \\
				4            & --          & 2                  & 0             & 2                  & --                                      & 10.12         & 7.50            & 7.5           & 11.15      \\
				5            & 1           & 4                  & 0             & 1                  & --                                      & 5.42          & 6.04            & 3.75          & 4.63       \\
				5            & 2           & 4                  & 0             & 1                  & --                                      & 5.42          & 6.04            & 3.75          & 4.67       \\
				6            & 3           & 4                  & 0             & 1                  & --                                      & 5.38          & 6.04            & 3.75          & 1.46       \\
				6            & 4           & 4                  & 0             & 1                  & --                                      & 5.33          & 6.04            & 3.75          & 4.62       \\
				7            & 5           & 7                  & 37.12         & 1                  & N(5,(1)\textasciicircum{}2)             & 6.54          & 8.58            & 10.69         & 6.81       \\
				7            & 6           & 7                  & 37.12         & 1                  & N(5,(1)\textasciicircum{}2)             & 6.59          & 8.58            & 10.69         & 6.99       \\
				&             &                    &               & \multicolumn{2}{l}{Cost} & 103.77       & 101.59            & 101.47        & 482.63       \\
				&             &                    &               &                    &                                         &               &                    &               &               \\
				Instance 3   &             &                    &               &                    &                                         &               &                    &               &               \\
				&             &                    &               &                    &                                         &               &                    &               &               \\
				&             &                    &               &                    &                                         & \multicolumn{4}{c}{OUL}  \\
				Node         & Predecessor & H & S & LT          & Demand                                  & DNNSMIO       & CD & Enumeration   & DFO           \\
				1            & --          & 0.4                & 0             & 2                  & --                                      & 40.98         & 36.67            & 35.55       & 39.66      \\
				2            & --          & 0.4                & 0             & 2                  & --                                      & 40.99         & 36.67            & 35.55       & 22.78      \\
				3            & --          & 0.4                & 0             & 2                  & --                                      & 41.03         & 36.67            & 35.55       & 12.47      \\
				4            & --          & 0.4                & 0             & 2                  & --                                      & 41.07         & 36.67            & 35.55       & 46.63      \\
				5            & 1           & 0.9                & 0             & 2                  & --                                      & 42.83         & 45.19            & 41.11       & 39.61       \\
				5            & 2           & 0.9                & 0             & 2                  & --                                      & 43.29         & 45.19            & 41.11       & 42.00      \\
				6            & 3           & 0.9                & 0             & 2                  & --                                      & 43.26         & 45.19            & 41.11       & 42.41       \\
				6            & 4           & 0.9                & 0             & 2                  & --                                      & 42.17         & 45.19            & 41.11       & 40.04      \\
				7            & 5           & 2.1                & 15            & 2                  & N(20,(3)\textasciicircum{}2)            & 46.19         & 47.85            & 52.22       & 47.02      \\
				7            & 6           & 2.1                & 15            & 2                  & N(20,(3)\textasciicircum{}2)            & 46.14         & 47.85            & 52.22       & 48.82      \\
				&             &                    &               & \multicolumn{2}{l}{Cost} & 163.15       & 161.30            & 161.13      & 441.43       \\
				&             &                    &               &                    &                                         &               &                    &               &               \\
				Instance 4   &             &                    &               &                    &                                         &               &                    &               &               \\
				&             &                    &               &                    &                                         &               &                    &               &               \\
				&             &                    &               &                    &                                         & \multicolumn{4}{c}{OUL}  \\
				Node         & Predecessor & H & S & LT          & Demand                                  & DNNSMIO       & CD & Enumeration   & DFO           \\
				1            & --          & 0.3                & 0             & 2                  & --                                      & 10.12         & 9.21            & 10.27       & 1.26       \\
				2            & --          & 0.3                & 0             & 2                  & --                                      & 10.19         & 9.21            & 10.2        & 7.59       \\
				3            & --          & 0.3                & 0             & 2                  & --                                      & 10.73         & 9.21            & 10.27       & 9.89       \\
				4            & --          & 0.3                & 0             & 2                  & --                                      & 10.21         & 9.21            & 10.27       & 1.60       \\
				5            & 1           & 0.8                & 0             & 2                  & --                                      & 12.74         & 11.55           & 10.27       & 8.06       \\
				5            & 2           & 0.8                & 0             & 2                  & --                                      & 12.38         & 11.55           & 10.27       & 7.26       \\
				6            & 3           & 0.8                & 0             & 2                  & --                                      & 12.55         & 11.55           & 10.27       & 5.27       \\
				6            & 4           & 0.8                & 0             & 2                  & --                                      & 12.20         & 11.55           & 10.27       & 7.94       \\
				7            & 5           & 2                  & 15            & 2                  & N(5,(1)\textasciicircum{}2)             & 12.25         & 12.62           & 13.05       & 15.52      \\
				7            & 6           & 2                  & 15            & 2                  & N(5,(1)\textasciicircum{}2)             & 12.25         & 12.62           & 13.05       & 13.81       \\
				&             &                    &               & \multicolumn{2}{l}{Cost} & 37.49        & 35.97              & 35.98      & 139.77       \\
				&             &                    &               &                    &                                         &               &                    &               &               \\
				Instance 5   &             &                    &               &                    &                                         &               &                    &               &               \\
				&             &                    &               &                    &                                         & \multicolumn{4}{c}{OUL}  \\
				Node         & Predecessor & H & S & LT          & Demand                                  & DNNSMIO       & CD & Enumeration   & DFO           \\
				1            & --          & 0.5                & 0             & 1                  & --                                      & 5.01          & 4.38            & 3.75       & 8.79       \\
				2            & --          & 0.5                & 0             & 1                  & --                                      & 5.02          & 4.38            & 3.75       & 7.35       \\
				3            & --          & 0.5                & 0             & 1                  & --                                      & 5.11          & 4.38            & 3.75       & 8.91       \\
				4            & --          & 0.5                & 0             & 1                  & --                                      & 5.06          & 4.38            & 3.75       & 5.23       \\
				5            & 1           & 1                  & 0             & 1                  & --                                      & 7.04          & 6.23            & 6.52       & 3.07       \\
				5            & 2           & 1                  & 0             & 1                  & --                                      & 7.07          & 6.23            & 6.52       & 5.71       \\
				6            & 3           & 1                  & 0             & 1                  & --                                      & 6.98          & 6.23            & 6.52       & 3.08       \\
				6            & 4           & 1                  & 0             & 1                  & --                                      & 6.91          & 6.23            & 6.52       & 5.74       \\
				7            & 5           & 3                  & 16            & 1                  & N(5,(1)\textasciicircum{}2)             & 6.33          & 6.49            & 6.52       & 5.75       \\
				7            & 6           & 3                  & 16            & 1                  & N(5,(1)\textasciicircum{}2)             & 6.23          & 6.49            & 6.52       & 6.38       \\
				&             &                    &               & \multicolumn{2}{l}{Cost} & 29.04      & 27.53         & 27.45       & 36.03       \\
				&             &                    &               &                    &                                         &               &                    &               &              
			\end{tabular}
		\end{scriptsize}
	\end{table*}

	\begin{table*}[]
		\caption{Assembly(2) SCN cases details}
		\label{tabassem2}
		\setlength\tabcolsep{4pt}
		\centering
		\begin{scriptsize}
			\begin{tabular}{llllllllll}
				Instance   1 &             &      &      &    &                              &         &         &             &          \\
				&             &      &      &    &                              & \multicolumn{4}{c}{Suggested OUL}          \\
				Node         & Predecessor & H    & S    & LT & Demand                       & DNNSMIO & CD      & Enumeration & DFO      \\
				1            & --          & 2    & 0    & 1  & --                           & 5.23    & 3.97 & 3.75        & 6.76  \\
				2            & --          & 2    & 0    & 1  & --                           & 5.01    & 3.97 & 3.75        & 6.53  \\
				3            & --          & 2    & 0    & 1  & --                           & 5.22    & 3.97 & 3.75        & 6.80  \\
				4            & --          & 2    & 0    & 1  & --                           & 4.96    & 3.97 & 3.75        & 6.80  \\
				5            & --          & 2    & 0    & 1  & --                           & 4.92    & 3.97 & 3.75        & 5.93  \\
				6            & 4           & 4    & 0    & 1  & --                           & 5.9     & 5.49 & 5.83     & 0.80  \\
				6            & 5           & 4    & 0    & 1  & --                           & 6.31    & 5.49 & 5.83     & 5.92  \\
				7            & 1           & 7    & 40   & 1  & N(5,(1)\textasciicircum{}2)  & 5.89    & 7.77 & 7.91     & 6.64  \\
				7            & 2           & 7    & 40   & 1  & N(5,(1)\textasciicircum{}2)  & 6.97    & 7.77 & 7.91     & 5.75  \\
				7            & 3           & 7    & 40   & 1  & N(5,(1)\textasciicircum{}2)  & 5.82    & 7.77 & 7.91     & 6.10   \\
				7            & 6           & 7    & 40   & \multicolumn{2}{l}{1}             & 6.33    & 7.77 & 7.91     & 6.27  \\
				&             &      &      &    & Cost                         & 93.9432 & 90.4087 & 90.5432     & 116.38   \\
				&             &      &      &    &                              &         &         &             &          \\
				Instance 2   &             &      &      &    &                              &         &         &             &          \\
				&             &      &      &    &                              & \multicolumn{4}{c}{Suggested OUL}          \\
				Node         & Predecessor & H    & S    & LT & Demand                       & DNNSMIO & CD      & Enumeration & DFO      \\
				1            & --          & 0.3  & 0    & 1  & --                           & 9.93    & 8.98 & 7.5         & 10.89 \\
				2            & --          & 0.3  & 0    & 1  & --                           & 9.89    & 8.98 & 7.5         & 10.49 \\
				3            & --          & 0.3  & 0    & 1  & --                           & 9.98    & 8.98 & 7.5         & 11.01 \\
				4            & --          & 0.3  & 0    & 1  & --                           & 9.86    & 8.98 & 7.5         & 11.24  \\
				5            & --          & 0.3  & 0    & 1  & --                           & 9.87    & 8.98 & 7.5         & 10.79  \\
				6            & 4           & 0.5  & 0    & 1  & --                           & 10.99   & 10.53 & 10.27     & 5.59   \\
				6            & 5           & 0.5  & 0    & 1  & --                           & 11.39   & 10.53 & 10.27     & 10.47 \\
				7            & 1           & 0.9  & 3.5  & 1  & N(10,(1)\textasciicircum{}2) & 12.47   & 12.54 & 14.44     & 10.88 \\
				7            & 2           & 0.9  & 3.5  & 1  & N(10,(1)\textasciicircum{}2) & 12.31   & 12.54 & 14.44     & 11.25 \\
				7            & 3           & 0.9  & 3.5  & 1  & N(10,(1)\textasciicircum{}2) & 12.5    & 12.54 & 14.44     & 11.12 \\
				7            & 6           & 0.9  & 3.5  & 1  & N(10,(1)\textasciicircum{}2) & 11.85   & 12.54 & 14.44     & 11.28 \\
				&             &      &      &    & Cost                         & 23.00 & 22.43 & 22.48     & 25.75  \\
				&             &      &      &    &                              &         &         &             &          \\
				Instance 3   &             &      &      & \multicolumn{2}{l}{}              &         &         &             &          \\
				&             &      &      &    &                              & \multicolumn{4}{c}{Suggested OUL}          \\
				Node         & Predecessor & H    & S    & LT & Demand                       & DNNSMIO & CD      & Enumeration & DFO      \\
				1            & --          & 0.5  & 0    & 1  & --                           & 10.7    & 11.1868 & 10.27     & 11.57 \\
				2            & --          & 0.5  & 0    & 1  & --                           & 10.43   & 11.1868 & 10.27     & 19.27 \\
				3            & --          & 0.5  & 0    & 1  & --                           & 10.71   & 11.1868 & 10.27     & 11.59 \\
				4            & --          & 0.5  & 0    & 1  & --                           & 10.27   & 11.1868 & 10.27     & 13.77 \\
				5            & --          & 0.5  & 0    & 1  & --                           & 10.24   & 11.1868 & 10.27     & 13.63 \\
				6            & 4           & 3    & 0    & 1  & --                           & 11.06   & 11.1868 & 10.27     & 5.71  \\
				6            & 5           & 3    & 0    & 1  & --                           & 11.07   & 11.1868 & 10.27     & 11.56  \\
				7            & 1           & 6    & 25   & 1  & N(10,(2)\textasciicircum{}2) & 11.54   & 12.4788 & 13.05     & 11.81 \\
				7            & 2           & 6    & 25   & 1  & N(10,(2)\textasciicircum{}2) & 11.94   & 12.4788 & 13.05     & 11.78 \\
				7            & 3           & 6    & 25   & 1  & N(10,(2)\textasciicircum{}2) & 10.87   & 12.4788 & 13.05     & 11.77 \\
				7            & 6           & 6    & 25   & 1  & N(10,(2)\textasciicircum{}2) & 11.62   & 12.4788 & 13.05     & 11.90 \\
				&             &      &      &    & Cost                         & 86.61 & 82.67 & 82.32     & 90.71  \\
				Instance 4   &             &      &      &    &                              &         &         &             &          \\
				&             &      &      &    &                              & \multicolumn{4}{c}{Suggested OUL}          \\
				Node         & Predecessor & H    & S    & \multicolumn{2}{l}{LT}            & DNNSMIO & CD      & Enumeration & DFO      \\
				1            & --          & 0.6  & 0    & 1  & --                           & 7.26    & 6.2706  & 5.25        & 7.47  \\
				2            & --          & 0.6  & 0    & 1  & --                           & 7.16    & 6.2706  & 5.25        & 7.95  \\
				3            & --          & 0.6  & 0    & 1  & --                           & 7.1     & 6.2706  & 5.25        & 6.10  \\
				4            & --          & 0.6  & 0    & 1  & --                           & 6.93    & 6.2706  & 5.25        & 5.57  \\
				5            & --          & 0.6  & 0    & 1  & --                           & 6.88    & 6.2706  & 5.25        & 7.52  \\
				6            & 4           & 1.1  & 0    & 1  & --                           & 7.72    & 7.53 & 7.19     & 10.76 \\
				6            & 5           & 1.1  & 0    & 1  & --                           & 8.02    & 7.53 & 7.19     & 5.42  \\
				7            & 1           & 2.5  & 5.4  & 1  & N(7,(1)\textasciicircum{}2)  & 8.22    & 8.72 & 10.11     & 7.60  \\
				7            & 2           & 2.5  & 5.4  & 1  & N(7,(1)\textasciicircum{}2)  & 7.87    & 8.72 & 10.11     & 7.87   \\
				7            & 3           & 2.5  & 5.4  & 1  & N(7,(1)\textasciicircum{}2)  & 7.73    & 8.72 & 10.11     & 8.18   \\
				7            & 6           & 2.5  & 5.4  & 1  & N(7,(1)\textasciicircum{}2)  & 7.43    & 8.72 & 10.11     & 9.30   \\
				&             &      &      &    & Cost                         & 34.62 & 34.04 & 34.17     & 42.35   \\
				&             &      &      &    &                              &         &         &             &          \\
				Instance 5   &             &      &      &    &                              &         &         &             &          \\
				&             &      &      &    &                              & \multicolumn{4}{c}{Suggested OUL}          \\
				Node         & Predecessor & H    & S    & LT & Demand                       & DNNSMIO & CD      & Enumeration & DFO      \\
				1            & --          & 0.25 & 0    & \multicolumn{2}{l}{1}             & 12.11   & 10.55 & 8.25        & 14.88 \\
				2            & --          & 0.25 & 0    & 1  & --                           & 11.97   & 10.55 & 8.25        & 14.55 \\
				3            & --          & 0.25 & 0    & 1  & --                           & 11.9    & 10.55 & 8.25        & 14.48 \\
				4            & --          & 0.25 & 0    & 1  & --                           & 11.5    & 10.55 & 8.25        & 4.95  \\
				5            & --          & 0.25 & 0    & 1  & --                           & 11.59   & 10.55 & 8.25        & 14.22 \\
				6            & 4           & 0.59 & 0    & 1  & --                           & 13.52   & 12.62 & 12.83     & 2.07  \\
				6            & 5           & 0.59 & 0    & 1  & --                           & 13.59   & 12.62 & 12.83     & 13.95 \\
				7            & 1           & 0.88 & 49.7 & 1  & N(11,(2)\textasciicircum{}2) & 14.86   & 18.44 & 20.47     & 15.89    \\
				7            & 2           & 0.88 & 49.7 & 1  & N(11,(2)\textasciicircum{}2) & 14.49   & 18.44 & 20.47     & 15.05  \\
				7            & 3           & 0.88 & 49.7 & 1  & N(11,(2)\textasciicircum{}2) & 14.51   & 18.44 & 20.47     & 14.55  \\
				7            & 6           & 0.88 & 49.7 & 1  & N(11,(2)\textasciicircum{}2) & 14.29   & 18.44 & 20.47     & 17.67 \\
				&             &      &      &    & Cost                         & 30.98 & 28.19 & 27.96     & 62.76 
			\end{tabular}
		\end{scriptsize}
	\end{table*}
	
	\begin{figure*}
	\hspace*{-2.5cm}
	\scalebox{0.45}{\input{CaseStudyResults2.pgf}}
	\caption{Total cost and OUL decisions for the case study SCN.}
	\label{fig:CaseStudyResult}
\end{figure*}

	\section{Complex SCN Learning Curve and Extra Discussion}

	Figure \ref{fig:CaseStudyResult} illustrates the learning process and the optimal values for all 13 OULs to be optimized. The jumps in the subplots are related to the restarting procedure discussed in the previous section. We stop the procedure simply when there is no gain anymore in terms of objective function values. The best cost values found before restarting the training procedures were 626.17, 479.68, and 478.61. Therefore, we stop restarting the training procedure after two retrainings. Figure \ref{fig:CaseStudyResult} provides the following insights:
	
	\begin{itemize}
		\item When the cost structures are more complex such as nonlinear (potentially very large) shortage penalties, the optimal solutions are biased towards ordering extra items in the upstream echelons. DNN-SMEIO sets the $(0\leftarrow1)$ OUL to $101.44$, while the average lead-time demand that node $1$ realizes is only $3\times 15.08=45.24$. The difference between the two values is considerable, and if a practitioner without this knowledge uses enumeration or an {\em ad hoc} approach, it is probable that the best value found by the DNN-SMEIO method would not even be in the interval selected by the practitioner.
		\item In consensus with what we observed in the previous examples, the upstream echelons learn their OUL values more quickly than those downstream.      
	\end{itemize}

	\section{Complex SCN Cost Comparison}
	Table \ref{tabCostComparison} shows the structure details of the different experiments done for the Complex SCN studied in Section \textcolor{blue}{4.5} of the main paper as well as suggested OULs by DNNSMIO and DFO. H, local holding cost coefficients, stockout cost coefficients and salvage reward coefficients
	
	\begin{table*}[]
		\caption{Cost comparison cases: structure details and suggested OULs}
		\label{tabCostComparison}
		\hspace*{-1.5cm}
		\setlength\tabcolsep{1.5pt}
		\centering	
		\begin{scriptsize}

			\begin{tabular}{lllllllll}
				Instance   1 &             &                                 &                            &                            &           &                               &               &               \\
				&             &                                 &                            &                            &           &                               & Suggested OUL & Suggested OUL \\
				Node         & Predecessor & H & S & R & LT & Demand                        & DNNSMIO       & DFO           \\
				1            & --          & 2                               & 4                          & 1.25                       & 1         & --                            & 49.25         & 52.22203      \\
				2            & 1           & 4-3                             & 12-4                       & 1.5                        & 1         & --                            & 17.75         & 18.48301      \\
				3            & 1           & 4-3                             & 12-4                       & 1.5                        & 1         & --                            & 17.45         & 18.42489      \\
				4            & 1           & 4-3                             & 12-4                       & 1.5                        & 1         & --                            & 17.36         & 18.40821      \\
				5            & 2           & 7-6                             & 36-12                      & 1.25                       & 1         & --                            & 7.3           & 5.34762       \\
				5            & 3           & 7-6                             & 36-12                      & 1.25                       & 1         & --                            & 6.7           & 5.29257       \\
				5            & 4           & 7-6                             & 36-12                      & 1.25                       & 1         & --                            & 6.79          & 5.33311       \\
				6            & 2           & 7-6                             & 36-12                      & 1.25                       & 1         & --                            & 7.55          & 5.32846       \\
				6            & 3           & 7-6                             & 36-12                      & 1.25                       & 1         & --                            & 6.8           & 5.31856       \\
				6            & 4           & 7-6                             & 36-12                      & 1.25                       & 1         & --                            & 6.99          & 5.30957       \\
				7            & 2           & 7-6                             & 36-12                      & 1.25                       & 1         & N(5,(1)\textasciicircum{}2)   & 7.92          & 5.30319       \\
				7            & 3           & 7-6                             & 36-12                      & 1.25                       & 1         & N(5,(1)\textasciicircum{}2)   & 6.64          & 5.30318       \\
				7            & 4           & 7-6                             & 36-12                      & 1.25                       & 1         & N(5,(1)\textasciicircum{}2)   & 7.21          & 5.28193       \\
				&             &                                 &                            &                            &           & cost                          & 380.95        & 402.41        \\
				&             &                                 &                            &                            &           &                               &               &               \\
				Instance 2   &             &                                 &                            &                            &           &                               &               &               \\
				&             &                                 &                            &                            &           &                               & Suggested OUL & Suggested OUL \\
				Node         & Predecessor & H & S & R & LT & Demand                        & DNNSMIO       & DFO           \\
				1            & --          & 2                               & 4                          & 2                          & 1         & --                            & 56.53         & 57.33026      \\
				2            & 1           & 4-3                             & 12-4                       & 2                          & 1         & --                            & 18.64         & 20.50994      \\
				3            & 1           & 4-3                             & 12-4                       & 2                          & 1         & --                            & 19.96         & 18.88558      \\
				4            & 1           & 4-3                             & 12-4                       & 2                          & 1         & --                            & 19.07         & 19.12507      \\
				5            & 2           & 7-6                             & 36-12                      & 2                          & 1         & --                            & 7.8           & 7.24254       \\
				5            & 3           & 7-6                             & 36-12                      & 2                          & 1         & --                            & 6.91          & 4.22521       \\
				5            & 4           & 7-6                             & 36-12                      & 2                          & 1         & --                            & 12.21         & 6.3352        \\
				6            & 2           & 7-6                             & 36-12                      & 2                          & 1         & --                            & 8.36          & 6.94252       \\
				6            & 3           & 7-6                             & 36-12                      & 2                          & 1         & --                            & 7.56          & 4.40232       \\
				6            & 4           & 7-6                             & 36-12                      & 2                          & 1         & --                            & 11.83         & 7.46706       \\
				7            & 2           & 7-6                             & 36-12                      & 2                          & 1         & N(6,(1)\textasciicircum{}2)   & 8.19          & 7.31975       \\
				7            & 3           & 7-6                             & 36-12                      & 2                          & 1         & N(4,(1)\textasciicircum{}2)   & 5.97          & 4.39467       \\
				7            & 4           & 7-6                             & 36-12                      & 2                          & 1         & N(6,(2)\textasciicircum{}2)   & 11.8          & 7.27793       \\
				&             &                                 &                            &                            &           & cost                          & 419.13        & 442.42        \\
				&             &                                 &                            &                            &           &                               &               &               \\
				Instance 3   &             &                                 &                            &                            &           &                               &               &               \\
				&             &                                 &                            &                            &           &                               & Suggested OUL & Suggested OUL \\
				Node         & Predecessor & H & S & R & LT & Demand                        & DNNSMIO       & DFO           \\
				1            & --          & 2                               & 4                          & 1                          & 1         & --                            & 53.73         & 56.46892      \\
				2            & 1           & 4-3                             & 12-4                       & 2                          & 1         & --                            & 18.87         & 19.30803      \\
				3            & 1           & 4-3                             & 12-4                       & 2                          & 1         & --                            & 19.43         & 19.16313      \\
				4            & 1           & 4-3                             & 12-4                       & 2                          & 1         & --                            & 18.75         & 19.61025      \\
				5            & 2           & 7-6                             & 36-12                      & 1                          & 1         & --                            & 6.67          & 5.41518       \\
				5            & 3           & 7-6                             & 36-12                      & 1                          & 1         & --                            & 7.28          & 5.61027       \\
				5            & 4           & 7-6                             & 36-12                      & 1                          & 1         & --                            & 12.64         & 7.17493       \\
				6            & 2           & 7-6                             & 36-12                      & 1                          & 1         & --                            & 7.49          & 5.40836       \\
				6            & 3           & 7-6                             & 36-12                      & 1                          & 1         & --                            & 8.19          & 5.44619       \\
				6            & 4           & 7-6                             & 36-12                      & 1                          & 1         & --                            & 10.67         & 6.90065       \\
				7            & 2           & 7-6                             & 36-12                      & 1                          & 1         & N(5,(1)\textasciicircum{}2)   & 7.55          & 5.90935       \\
				7            & 3           & 7-6                             & 36-12                      & 1                          & 1         & N(5,(1.2)\textasciicircum{}2) & 8.59          & 4.75065       \\
				7            & 4           & 7-6                             & 36-12                      & 1                          & 1         & N(6,(1.5)\textasciicircum{}2) & 9.25          & 7.14491       \\
				&             &                                 &                            &                            &           & cost                          & 407.83        & 408.27        \\
				&             &                                 &                            &                            &           &                               &               &               \\
				Instance 4   &             &                                 &                            &                            &           &                               &               &               \\
				&             &                                 &                            &                            &           &                               & Suggested OUL & Suggested OUL \\
				Node         & Predecessor & H & S & R & LT & Demand                        & DNNSMIO       & DFO           \\
				1            & --          & 2                               & 4                          & 2.5                        & 1         & --                            & 48.56         & 52.62784      \\
				2            & 1           & 4-3                             & 12-4                       & 2.5                        & 1         & --                            & 17.56         & 17.88147      \\
				3            & 1           & 4-3                             & 12-4                       & 2.5                        & 1         & --                            & 16.9          & 17.9517       \\
				4            & 1           & 4-3                             & 12-4                       & 2.5                        & 1         & --                            & 17.62         & 16.90722      \\
				5            & 2           & 7-6                             & 36-12                      & 2.5                        & 1         & --                            & 7.11          & 5.15377       \\
				5            & 3           & 7-6                             & 36-12                      & 2.5                        & 1         & --                            & 7.85          & 7.34585       \\
				5            & 4           & 7-6                             & 36-12                      & 2.5                        & 1         & --                            & 7.22          & 4.16108       \\
				6            & 2           & 7-6                             & 36-12                      & 2.5                        & 1         & --                            & 7.19          & 5.48897       \\
				6            & 3           & 7-6                             & 36-12                      & 2.5                        & 1         & --                            & 8.39          & 7.18886       \\
				6            & 4           & 7-6                             & 36-12                      & 2.5                        & 1         & --                            & 6.39          & 4.39452       \\
				7            & 2           & 7-6                             & 36-12                      & 2.5                        & 1         & N(5,(1)\textasciicircum{}2)   & 7.64          & 5.83787       \\
				7            & 3           & 7-6                             & 36-12                      & 2.5                        & 1         & N(6,(1)\textasciicircum{}2)   & 8.75          & 6.75788       \\
				7            & 4           & 7-6                             & 36-12                      & 2.5                        & 1         & N(4,(1.5)\textasciicircum{}2) & 6.25          & 4.72713       \\
				&             &                                 &                            &                            &           & cost                          & 379.31        & 400.047       \\
				&             &                                 &                            &                            &           &                               &               &               \\
				&             &                                 &                            &                            &           &                               &               &              
			\end{tabular}
		\end{scriptsize}
	\end{table*}

	\bibliography{RLSCPaper}
	\bibliographystyle{unsrtnat}